\documentclass[conference]{IEEEtran}
\pdfoutput=1

\usepackage{times}

\usepackage{graphicx} 
\usepackage{subfigure} 
\usepackage[numbers]{natbib}
\usepackage{multicol}
\usepackage{caption}
\usepackage{verbatim}
\usepackage{amsmath}
\usepackage{wrapfig,multirow}
\usepackage{array}
\usepackage{relsize}
\usepackage{caption}
\usepackage{algorithm}
\usepackage{algorithmic}
\usepackage{amssymb,amsmath}
\usepackage{url}
\newtheorem{thm}{Theorem}
\newtheorem{cor}[thm]{Corollary}
\usepackage{rotating}

\usepackage[numbers]{natbib}
\usepackage{graphicx}
\usepackage{subfigure}
\usepackage{multirow}
\usepackage{multicol}
\usepackage{times}
\usepackage[table]{xcolor}
\usepackage{verbatim}
\usepackage{caption}
\usepackage{array}
\usepackage{relsize}
\usepackage{amssymb,amsmath}
\usepackage{url}
\usepackage{rotating}
\usepackage{wrapfig,booktabs,blindtext}
\usepackage{color}

\usepackage{color}

\graphicspath{{fig/}}

\newcommand{\header}[1]{\noindent\textbf{#1}}

 \belowdisplayskip \abovedisplayskip
\renewcommand{\baselinestretch}{0.97}

 \newlength\savedwidth

\newlength{\sectionReduceTop}
\newlength{\sectionReduceBot}
\newlength{\subsectionReduceTop}
\newlength{\subsectionReduceBot}
\newlength{\abstractReduceTop}
\newlength{\abstractReduceBot}
\newlength{\captionReduceTop}
\newlength{\captionReduceBot}
\newlength{\subsubsectionReduceTop}
\newlength{\subsubsectionReduceBot}

\newlength{\horSkip}
\newlength{\verSkip}

\newlength{\figureHeight}
\newcolumntype{L}[1]{>{\raggedright\let\newline\\\arraybackslash\hspace{0pt}}m{#1}}
\title{Learning Preferences for Manipulation Tasks\\ from Online Coactive Feedback}
\begin{document}
\author{Ashesh Jain, Shikhar Sharma, Thorsten Joachims, Ashutosh
Saxena\\
Department of Computer Science, Cornell University, USA.\\
\{ashesh, ss2986, tj, asaxena\}@cs.cornell.edu}
\maketitle
\begin{abstract}
We consider the problem of learning preferences over trajectories for mobile manipulators 
such as personal robots and assembly line robots. 
The preferences we learn are more intricate 
than simple geometric constraints on trajectories; they are rather governed by the surrounding context 
of various objects and human interactions in the environment. 
We propose a coactive online 
learning framework for teaching preferences in contextually rich environments. 
The key novelty of our approach lies in the type of feedback
expected from the user: the human user does not need to demonstrate optimal trajectories 
as training data, but merely needs to iteratively provide trajectories that slightly
improve over the trajectory currently proposed by the system. We argue that this 
coactive preference feedback can be more easily elicited than
demonstrations of optimal trajectories.
Nevertheless, theoretical
regret bounds of our algorithm match the asymptotic rates of optimal trajectory
algorithms. 

We implement our algorithm on two high degree-of-freedom robots, PR2 and 
Baxter, and present three intuitive mechanisms for providing such incremental feedback. 
In our experimental evaluation we consider two context rich settings -- household chores 
and grocery store checkout -- and show that users are able to train the robot 
with just a few feedbacks (taking only a few minutes).\footnote{Parts of this
work has been published at NIPS and ISRR conferences~\citep{Jain13,Jain13b}. This journal
submission presents a consistent full paper, and also includes the proof of
regret bounds, more details of the robotic system, and a thorough related work.}

\end{abstract}

\section{Introduction}
\label{sec:1}
Recent advances in robotics have resulted in mobile manipulators with high degree 
of freedom (DoF) arms.
However, the use of high DoF arms has so far been largely successful only in structured
environments such as  manufacturing scenarios, where they 
perform repetitive motions (e.g., recent deployment of Baxter on assembly lines). 
One challenge in the deployment of these robots in unstructured environments (such as a grocery  
checkout counter or at our homes) is their lack of understanding of user preferences and 
thereby not producing desirable motions.  In this work we address the problem of learning preferences 
over trajectories 
for high DoF robots such as Baxter or PR2. 
We consider a variety of household chores for PR2 and grocery checkout tasks for Baxter.

A key problem for high DoF manipulators lies in identifying an appropriate trajectory for a task. 
An appropriate trajectory not only needs to be valid from a geometric point (i.e., feasible and 
obstacle-free, the criteria that most path planners focus on), but it also needs to satisfy the 
user's preferences. Such users' preferences over trajectories can be common
across users or they may vary between users, between tasks, and between 
the environments the trajectory is performed in. For example, a household robot should move a glass of water 
in an upright position without jerks while maintaining a safe distance from nearby electronic devices.
In another example, 
a robot checking out a knife at a grocery 
store should strictly move it at a safe 
distance from nearby humans. Furthermore, straight-line trajectories in Euclidean space may no 
longer be the preferred ones. For example, trajectories of
heavy items should not pass over fragile items but rather move around them. These preferences are 
often hard to describe and anticipate without knowing where and how the robot is deployed.
This makes it infeasible to manually encode them in existing path planners 
(e.g., Zucker et al.~\citep{Zucker13}, Sucan et al.~\citep{Ompl}, Schulman et
al.~\citep{Schulman13}) a priori.

In this work we propose an algorithm for learning user preferences over trajectories through 
interactive feedback from the user in a coactive learning setting~\citep{Shivaswamy12}.
In this setting the robot learns through iterations of user feedback. At each
iteration robot receives a task and it predicts a trajectory. The user responds
by slightly improving the trajectory but not necessarily revealing the optimal
trajectory. The robot use this feedback from user to improve its predictions for future
iterations.
Unlike in other learning settings, where a human first demonstrates optimal 
trajectories for a task to the robot~\citep{Argall09}, our learning model does not rely on the user's ability 
to demonstrate optimal trajectories a priori. Instead, our learning algorithm explicitly guides the learning 
process and merely requires the user to incrementally improve the robot's
trajectories, thereby learning preferences of user and not the expert. From these interactive improvements the robot 
learns a general model of the user's preferences in an online fashion. We realize this learning
algorithm on PR2 and Baxter robots, and also leverage robot-specific design to
allow users to easily give preference 
feedback.

Our experiments show that a robot trained using this approach can autonomously perform
new tasks and if need be, only a small number of interactions are sufficient to tune the robot to the new task. 
Since the user does \textit{not} have to demonstrate a (near) optimal trajectory to the robot,
the feedback is easier to provide and more widely applicable. Nevertheless, it leads 
to an online learning algorithm with provable regret bounds that decay at the
same rate as for optimal demonstration algorithms (eg. Ratliff et
al.~\citep{Ratliff07a}).

\begin{figure*}[t]
\centering
\includegraphics[width=0.4\textwidth,natwidth=640,natheight=428]{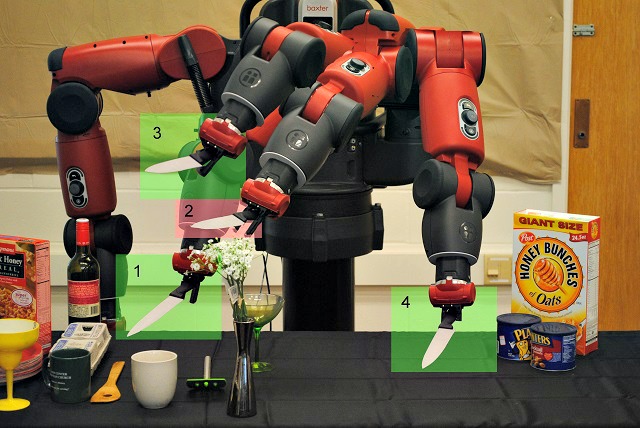}
\hspace*{5mm}
\includegraphics[width=0.4\textwidth,natwidth=1024,natheight=685]{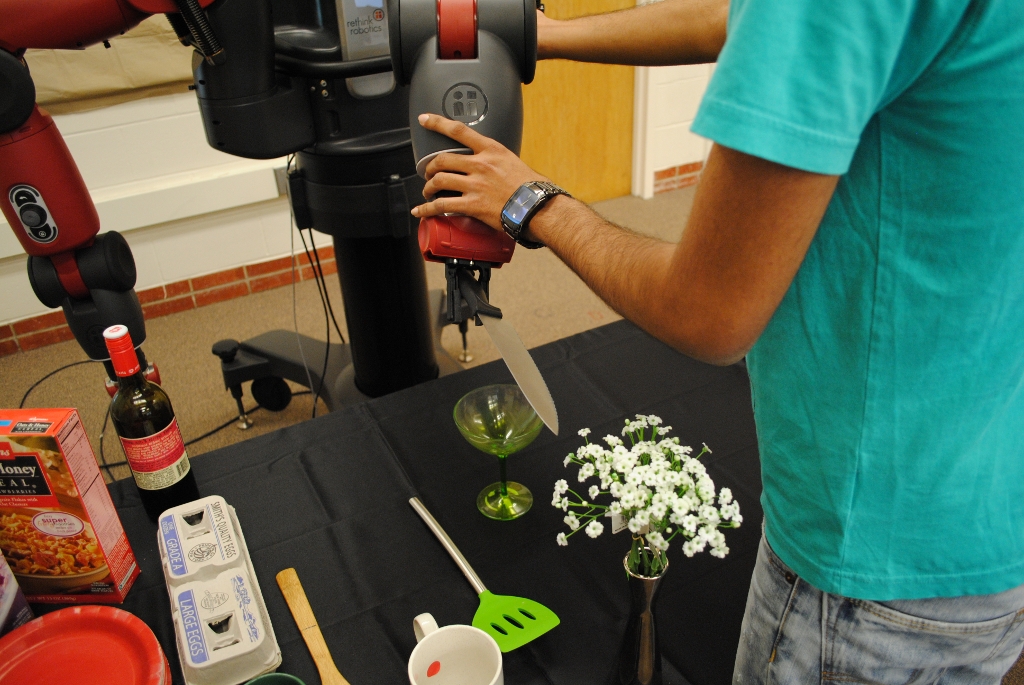}
\caption{\textbf{Zero-G feedback:} Learning trajectory preferences from suboptimal
\emph{zero-G} feedback. (\textbf{Left}) 
Robot plans a bad trajectory (waypoints 1-2-4) with knife close to flower. As feedback, user corrects waypoint 2 and moves it to waypoint 3. (\textbf{Right}) User providing \emph{zero-G} feedback
on waypoint 2.}
\label{fig:zeroG}
\end{figure*}

In our empirical evaluation we learn preferences for two high DoF robots, PR2 and Baxter, on a
variety of household and grocery checkout tasks respectively.
We design expressive trajectory features and show how our algorithm learns  
preferences from online user feedback on a broad range of tasks for which object properties are of 
particular importance (e.g., manipulating sharp objects with humans in the vicinity). 
We extensively evaluate our approach on a set of 35 household and 16 grocery checkout tasks, both
in batch experiments as well as through robotic experiments wherein users provide
their preferences to the robot.  Our results show that
our system not only quickly learns good trajectories on individual tasks,
but also generalizes well to tasks that the algorithm has not seen before. 
In summary, our key contributions are:
\begin{enumerate}
\item We present an approach for teaching robots which does not rely on experts'
demonstrations but nevertheless gives strong theoretical guarantees.
\item We design a robotic system with multiple easy to elicit
feedback mechanisms to improve the current trajectory. 
\item We implement our algorithm on two robotic platforms, PR2 and Baxter, and
support it with a user study. 
\item We consider preferences that go beyond simple geometric criteria to
capture object and human interactions.
\item We design expressive trajectory features to capture contextual
information. These features might also find use in other robotic applications. 
\end{enumerate}

In the following section we discuss related works. In section~\ref{sec:baxter} 
we describe our system and feedback mechanisms.
Our learning algorithm and trajectory features are discussed in
sections~\ref{sec:approach} and~\ref{sec:algorithm}, respectively. Section~\ref{sec:experiment}
gives our experiments and results. We discuss future research directions and
conclude in section~\ref{sec:conclusion}.

\section{Related Work}
\label{sec:related}

Path planning is one of the key problems in robotics. Here, the objective is to
find a collision free path from a start to goal location. Over the last decade
many planning algorithms have been proposed, such as sampling based planners by
Lavalle and Kuffner~\citep{Lavalle01}, and Karaman and Frazzoli~\citep{Karaman10},
search based planners by Cohen et al.~\citep{Sbpl}, trajectory optimizers by
Schulman et al.~\citep{Schulman13}, and Zucker et al.~\citep{Zucker13} and many
more~\citep{Karaman11}. However, given the large space of possible trajectories in most robotic applications 
simply a collision free trajectory might not suffice, instead the trajectory should
satisfy certain constraints and obey the end user preferences. Such preferences
are often encoded as a cost which planners optimize~\citep{Karaman10,Schulman13,Zucker13}. We address the problem of
learning a cost over trajectories for context-rich environments, and from
sub-optimal feedback elicited from non-expert users. We now describe related work
in various aspects of this problem. 

\noindent \textbf{Learning from Demonstration (LfD):} Teaching a robot to produce desired motions has been a long standing goal 
and several approaches have been studied.  In LfD an expert provides
demonstrations of optimal trajectories and the robot tries to mimic the expert.
Examples of LfD includes,
autonomous helicopter flights~\citep{Abbeel10}, ball-in-a-cup game~\citep{Kober11}, 
planning 2-D paths~\citep{Ratliff09b,Ratliff06}, etc.  Such settings  
assume that kinesthetic demonstrations are intuitive to an end-user and it is clear to an expert what constitutes 
a good trajectory. In many scenarios, especially involving high DoF manipulators, this is extremely challenging 
to do~\citep{Akgun12}.\footnote{Consider the following analogy. In search engine results, it is much 
harder for the user to provide the best web-pages for each query, but it is easier to provide relative ranking on the search results by clicking.}
This is because the users have to give not only the end-effector's location at each time-step, 
but also the full configuration of the arm in a spatially and temporally consistent manner.
In Ratliff et al.~\citep{Ratliff07a} the robot observes optimal user feedback but
performs approximate inference. On the other hand, in our setting, the user never discloses the optimal trajectory or feedback,
but instead, the robot learns preferences from sub-optimal suggestions for how
the trajectory can be improved. 
\\~\\~
\textbf{Noisy demonstrations and other forms of user feedback:} Some later works in LfD provided ways for handling noisy demonstrations, 
under the assumption that demonstrations are either near optimal, as in Ziebart
et al.~\citep{Ziebart08}, or locally 
optimal, as in Levine et al.~\citep{Levine12}. Providing noisy demonstrations is different from providing relative preferences, 
which are biased and can be far from optimal. We compare with an algorithm for noisy LfD learning in our experiments. 
Wilson et al.~\citep{Wilson12} proposed a Bayesian framework for learning
rewards of a Markov decision process via trajectory preference queries. 
Our approach advances over~\citep{Wilson12} and Calinon
et.~al.~\citep{Calinon07} in that we model user as a utility maximizing agent.
Further, our score function theoretically converges to user's hidden function despite recieving 
sub-optimal feedback.
\begin{figure*}[t]
\centering
\includegraphics[width=0.32\textwidth,height=0.23\textwidth,natwidth=1358,natheight=759]{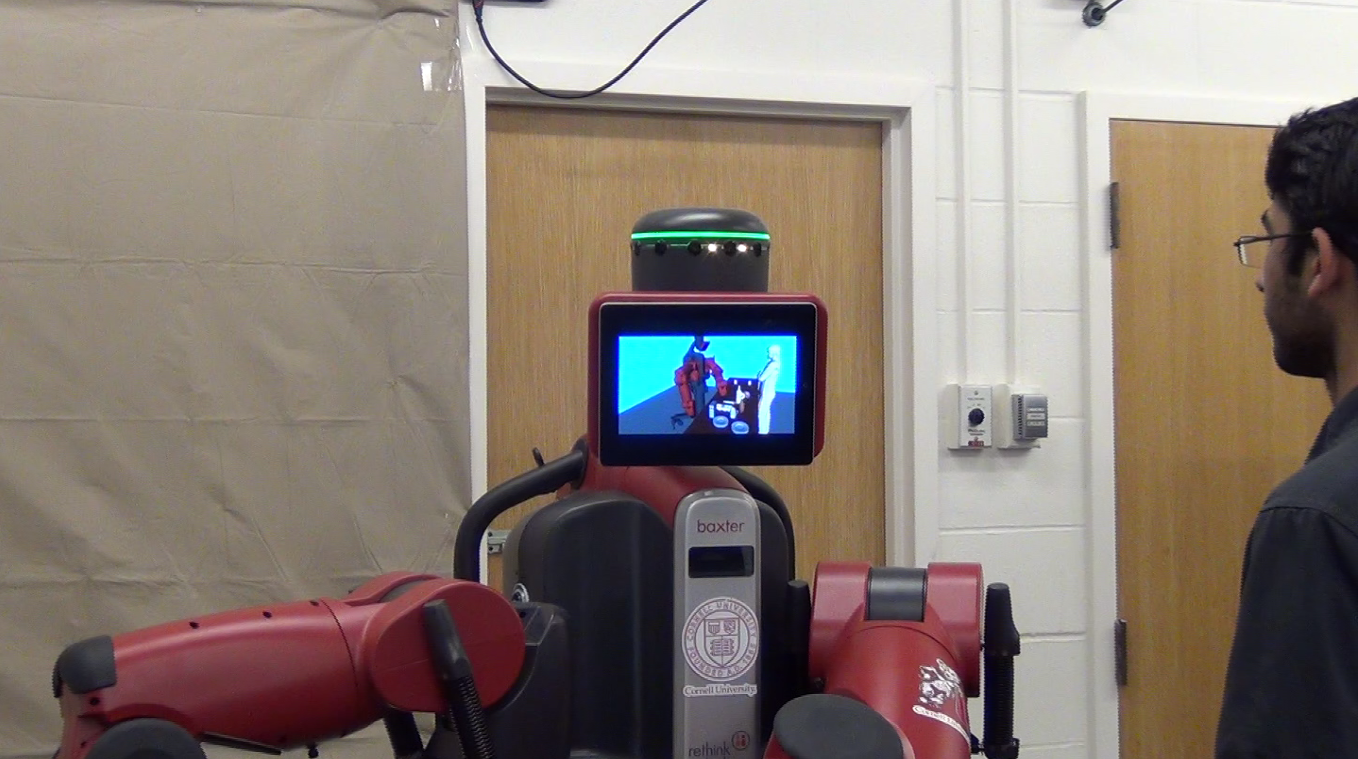}
\includegraphics[width=0.32\textwidth,height=0.23\textwidth,natwidth=1920,natheight=1080]{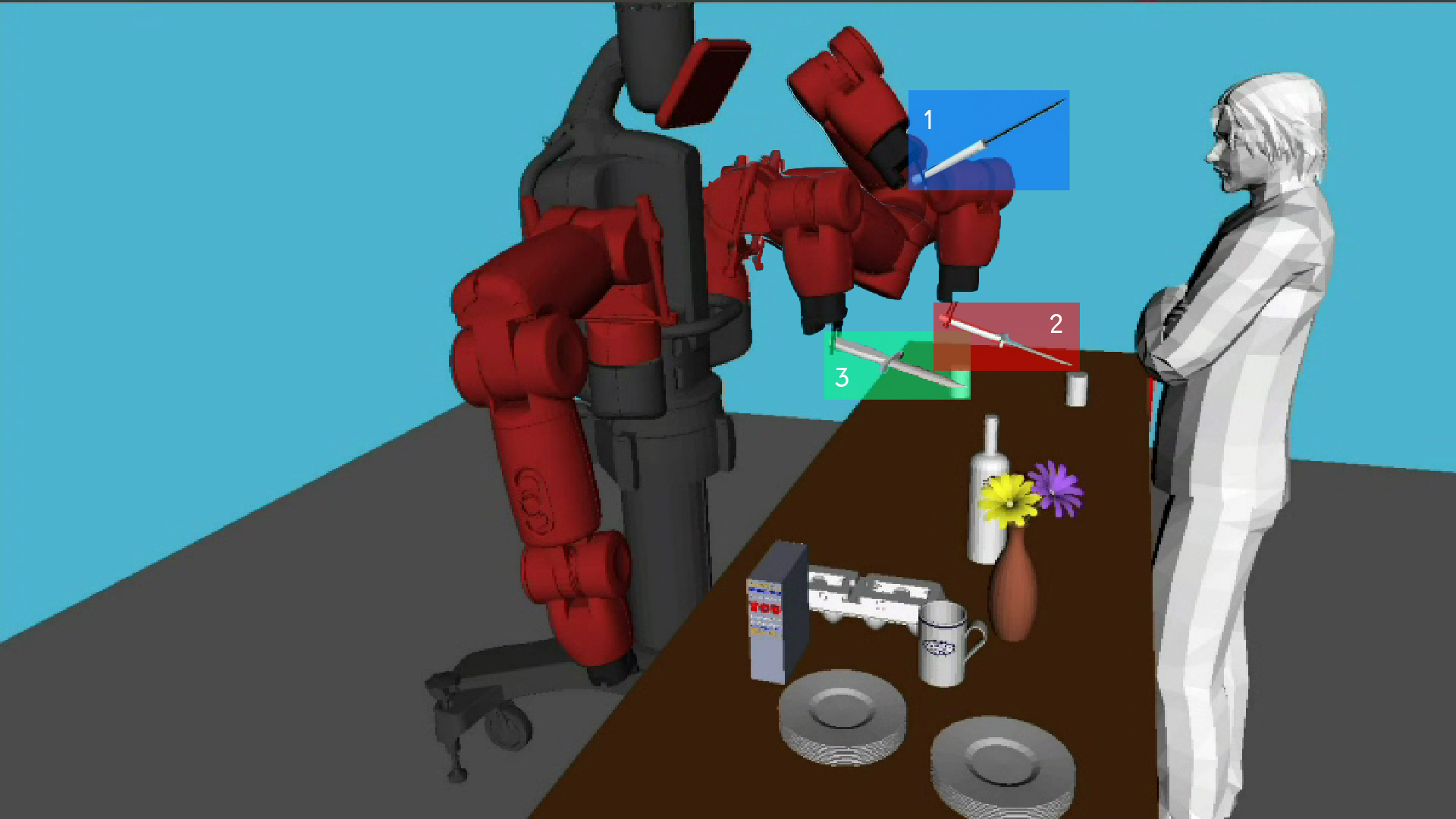}
\includegraphics[width=0.32\textwidth,height=0.23\textwidth,natwidth=1920,natheight=1080]{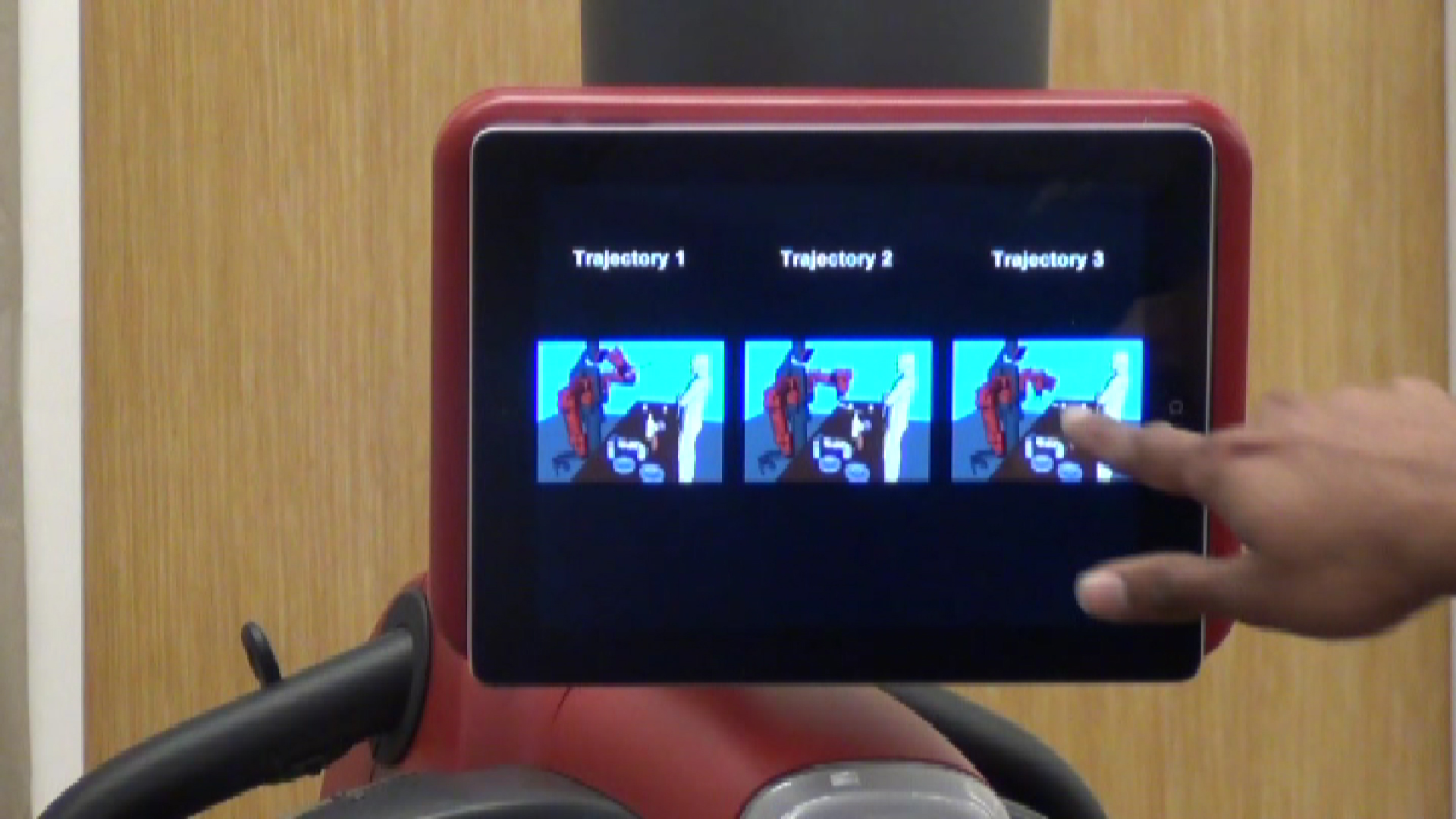}
\caption{\textbf{Re-rank feedback mechanism:} \textbf{(Left)} Robot ranks trajectories using the score function and 
\textbf{(Middle)} displays top three trajectories on a touch screen device (iPad here). \textbf{(Right)} As feedback, the user
improves the ranking by selecting the third trajectory.}
\label{fig:rerank_sim}
\end{figure*}
In the past, various interactive methods (e.g. human gestures)~\citep{Bischoff02,Stopp01} 
have been employed to teach assembly line robots.
However, these methods required the user to interactively show the complete
sequence of actions, which the robot then remembered for future use. 
Recent works by Nikolaidis et al.~\citep{Nikolaidis12,Nikolaidis13} in human-robot collaboration 
learns human preferences over a sequence of sub-tasks
in assembly line manufacturing. However, these works are agnostic 
to the user preferences over robot's trajectories.
Our work could complement theirs to achieve better human-robot collaboration.
\\~\\~
\noindent \textbf{Learning preferences over trajectories:} User preferences over
robot's trajectories have been studied in human-robot
interaction. Sisbot et.~al.~\citep{Sisbot07,Sisbot07b} and Mainprice et.~al.~\citep{Mainprice11}
planned trajectories satisfying user specified preferences in form of
constraints on the distance of the robot from the user, 
the visibility of the robot and the user's arm comfort. Dragan et.~al.~\citep{Dragan13b} used functional gradients~\citep{Chomp} 
to optimize for legibility of robot trajectories. We differ from these in that
we take a data driven approach and \textit{learn} score functions reflecting user preferences from sub-optimal feedback. 
\\~\\~
\noindent \textbf{Planning from a cost function:} In many applications, the goal is to find a trajectory that optimizes a cost function. 
Several works build upon the sampling-based planner RRT~\citep{Lavalle01} to optimize various cost heuristics~\citep{Ettlin06,Ferguson06,Jaillet10}. Additive cost functions with Lipschitz continuity can
be optimized using optimal planners such as RRT*~\citep{Karaman10}. Some
approaches introduce sampling bias~\citep{Leven03} to guide the sampling based planner. 
Recent trajectory optimizers such as CHOMP~\citep{Chomp} and
TrajOpt~\citep{Schulman13} provide optimization based approaches to finding optimal
trajectory.
Our work is complementary to these in that we learn a cost function while the
above approaches optimize a cost.

Our work is also complementary to few works in path planning. 
Berenson et al.~\citep{Berenson12} and Phillips et al.~\citep{Phillips12}
consider the problem of trajectories for high-dimensional manipulators. 
For computational reasons they create a database of prior trajectories, which we
could leverage to train our system. Other recent works
consider generating human-like
trajectories~\citep{Dragan13a,Dragan13b,Yamane13}. Humans-robot interaction is
an important aspect and our approach could incorporate similar ideas.
\\~\\~
\textbf{Application domain:} In addition to above mentioned differences we
also differ in the applications we address. We capture the necessary contextual
information for household and grocery store robots, while such context is absent
in previous works. 
Our application scenario of learning trajectories for high DoF 
manipulations performing 
tasks in presence of different objects and environmental constraints goes beyond the
application scenarios that previous works have considered. 
Some works in mobile robotics learn context-rich perception-driven cost
functions, such as Silver et al.~\citep{Silver10},
Kretzschmar et al.~\citep{Kretzschmar14} and Kitani et al.~\citep{Kitani12}.
In this work we use features that consider
robot configurations, object-object relations, and temporal behavior, and use
them to learn a score function representing the preferences over trajectories.

\section{Coactive learning with incremental feedback}
\label{sec:baxter}
We first give an overview of our robot learning setup and then describe
in detail three mechanisms of user feedback. 

\subsection{Robot learning setup}
We propose an online algorithm for learning preferences in trajectories from sub-optimal user feedback. 
At each step the robot receives a task as input
and outputs a trajectory that maximizes its current estimate of some score function. 
It then observes user feedback -- an improved trajectory --
and updates the score function to better match the user preferences.
This procedure of learning via iterative improvement is known as coactive
learning. We implement the algorithm on PR2 and Baxter robots, both having two 7
DoF arms. In the process of training, the initial trajectory
proposed by the robot can be far-off the desired behavior. Therefore, instead of
directly executing trajectories in human environments, users first visualize
them 
in the OpenRAVE simulator~\citep{Diankov10} and then decide the kind of feedback
they would like to provide. 

\subsection{Feedback mechanisms}
\begin{figure*}[t]
\centering
\begin{minipage}{.48\textwidth}
\centering
\includegraphics[width=\linewidth,height=0.6\linewidth,natwidth=640,natheight=428]{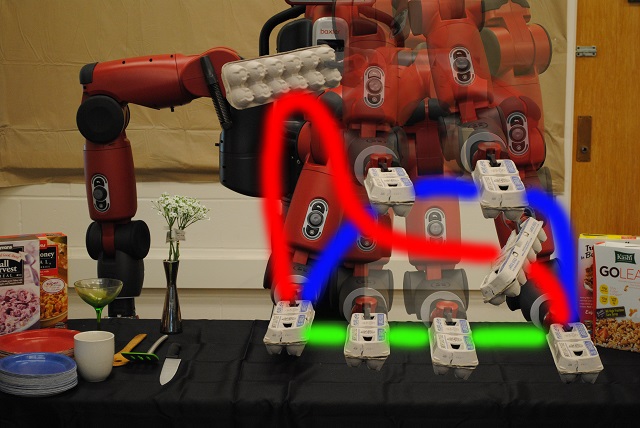}
\captionof{figure}{\textbf{Re-ranking feedback:} Shows three trajectories for moving egg carton from left to right.
Using the current estimate of score function robot ranks them  
as {\color{red}red}, {\color{green}green} and
{\color{blue}blue}. As feedback user clicks 
the {\color{green}green} trajectory. \textbf{Preference}: Eggs are fragile. 
They should be kept upright and near the supporting surface.}
\label{fig:rerank}
\end{minipage}%
\hskip 0.04\textwidth
\begin{minipage}{.48\textwidth}
\centering
\includegraphics[width=\linewidth,height=0.6\linewidth,natwidth=1300,natheight=600]{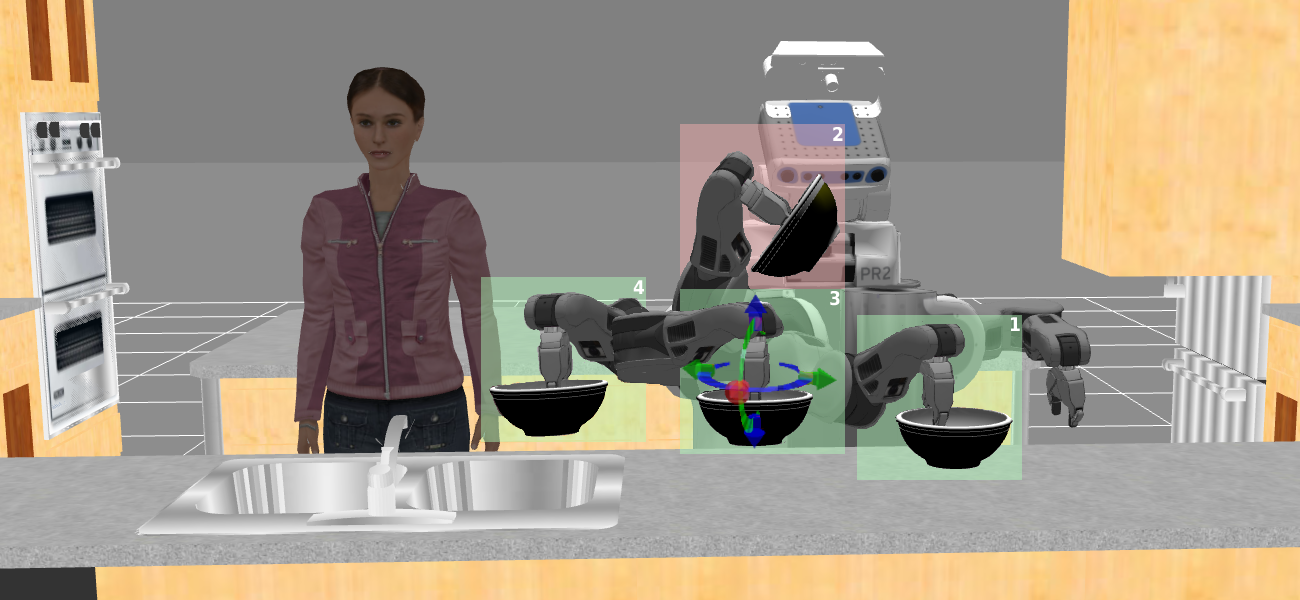}
\captionof{figure}{\textbf{Interactive feedback.} Task here is to move a bowl filled with water. The robot presents a bad 
trajectory with waypoints 1-2-4 to the user.  As feedback user moves waypoint 2 (red) to waypoint 3 (green) using Rviz interactive markers.
The interactive markers guides the user to correct the waypoint. }
\label{fig:rviz-ros}
\end{minipage}
\end{figure*}

Our goal is to learn even from feedback given by non-expert users.  
We therefore require the feedback only to be \emph{incrementally} better (as compared
to being close to optimal) in expectation, and will show that such feedback is sufficient for the algorithm's
convergence.
This stands in contrast to learning from demonstration (LfD) 
methods~\citep{Abbeel10,Kober11,Ratliff09b,Ratliff06} which require (near) optimal demonstrations of the complete trajectory. Such
demonstrations can be extremely challenging and non-intuitive to provide for many high DoF manipulators~\citep{Akgun12}.
Instead, we found~\citep{Jain13b,Jain13} that it is more intuitive for users to give incremental
feedback on high DoF arms by improving upon a proposed trajectory.
We now summarize three feedback mechanisms that enable the user to iteratively provide improved trajectories.
\\~\\~
\textit{(a) Re-ranking:} We display the ranking of trajectories using OpenRAVE~\citep{Diankov10}
 on a touch screen device and ask the user to identify whether any of the lower-ranked trajectories is better than the top-ranked one.
 The user sequentially observes the trajectories in order of their current predicted scores and clicks on the first trajectory
 which is better than the top ranked trajectory. Figure~\ref{fig:rerank_sim}
shows three trajectories for moving a knife. 
 As feedback, the user moves the trajectory at rank 3 to the top position. Likewise, Figure~\ref{fig:rerank} shows three trajectories for moving an egg carton. 
 Using the current estimate of score function robot ranks them as red
($1^{st}$), green ($2^{nd}$) and blue ($3^{rd}$). Since eggs are fragile the user 
 selects the green trajectory.
\\~\\~
\textit{(b) Zero-G:} This is a kinesthetic feedback. It allows the user to correct trajectory waypoints by physically changing
the robot's arm configuration as shown in Figure~\ref{fig:zeroG}.  High DoF
arms such as the Barrett WAM and Baxter have zero-force gravity-compensation (zero-G) mode,
under which the robot's arms become light and the users can effortlessly steer
them to a desired configuration. On Baxter, this zero-G mode is automatically activated
when a user holds the robot's wrist (see Figure~\ref{fig:zeroG}, right).
We use this zero-G mode as a feedback method for incrementally improving
the trajectory by correcting a waypoint.
This feedback is useful (i) for bootstrapping the robot,
(ii) for avoiding local maxima where the top trajectories in the ranked list are
all bad but ordered correctly, and (iii) when the user is satisfied with the top
ranked trajectory except for minor errors. 
\\~\\~
\textit{(c) Interactive:} For the robots whose hardware does not permit zero-G feedback, such as PR2, we built an alternative interactive Rviz-ROS~\citep{Rviz} interface for
allowing the users to improve the trajectories by waypoint correction. Figure~\ref{fig:rviz-ros} shows a robot moving a bowl with one bad waypoint (in red),
and the user provides a feedback by correcting it. This feedback serves the same purpose as zero-G.
\\~\\~
\noindent Note that, in all three kinds of feedback the user never reveals the optimal trajectory
to the algorithm, but only provides a slightly improved trajectory (in expectation).

\section{Learning and Feedback Model}
\label{sec:approach}

\begin{figure*}
\centering
\includegraphics[width=.8\linewidth,height=.27\linewidth,natwidth=1598,natheight=476]{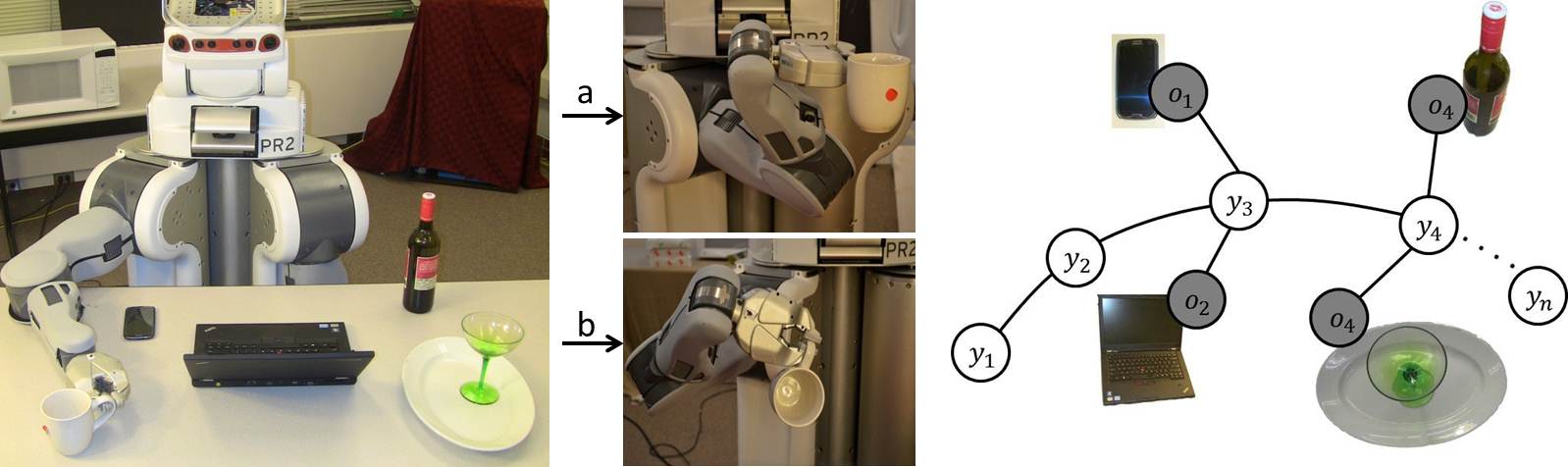}
\caption{\textbf{(Left)} An environment with a few objects where the robot was asked to move the cup on the left to 
the right. \textbf{(Middle)} There are two ways of moving it, `a' and `b', both are suboptimal in that the arm is contorted
in `a' but it tilts the cup in `b'. Given such constrained scenarios, we need to reason about such subtle 
preferences. \textbf{(Right)} We encode preferences concerned with object-object interactions in a score function expressed over a graph.
Here $y_1,\ldots,y_n$ are different waypoints in a trajectory.
The shaded nodes corresponds to environment (table node not shown here).
Edges denotes interaction between nodes.}
\label{fig:trajgraph}
\end{figure*}

We model the learning problem in the following way. For a given task, the robot is given a context $x$ that describes the environment, the objects, and any other input relevant to the
problem.  The robot has to figure out what is a good trajectory $y$ for this context.
Formally, we assume that the user has a scoring function $s^*(x,y)$ that reflects how much 
he values each trajectory $y$ for context $x$. The higher the score, the better the trajectory. 
Note that this scoring function cannot be 
observed directly, nor do we assume that the user can actually provide cardinal valuations 
according to this function. Instead, we merely assume that the user can provide us with 
{\em preferences} that reflect this scoring function. The robot's goal is to learn a function $s(x,y;w)$ (where $w$ are the parameters to be learned) that approximates the user's true scoring function $s^*(x,y)$ as closely as possible. 
\\~\\~
\noindent \textbf{Interaction Model.} The learning process proceeds through the following repeated cycle of interactions.
\begin{itemize}
\item \textbf{Step 1:} The robot receives a context $x$ and uses a planner to sample a set of trajectories, and ranks them according to its current approximate scoring function $s(x,y;w)$.
\item \textbf{Step 2:} The user either lets the robot execute the top-ranked trajectory, or 
corrects the robot by providing an improved trajectory $\bar{y}$. This provides 
feedback indicating that $s^*(x,\bar{y}) > s^*(x,y)$.
\item  \textbf{Step 3:}  The robot now updates the parameter $w$ of $s(x,y;w)$ based on this preference feedback and returns to step 1.
\end{itemize}
\textbf{Regret.} The robot's performance will be measured in terms of regret, 
$REG_T = \frac{1}{T} \sum_{t=1}^T [ s^*(x_t,{y}^*_t) - s^*(x_t,{y}_t) ]$,
which compares the robot's trajectory ${y}_t$ at each time step $t$ against the optimal 
trajectory ${y}_t^*$ maximizing the user's unknown scoring function $s^*(x,{y})$, 
${y}_t^* = argmax_{y} s^*(x_t,{y})$.
Note that the regret is expressed in terms of the user's true scoring function $s^*$, 
even though this function is \emph{never observed}. Regret characterizes 
the performance of the robot over its whole lifetime, therefore reflecting how well it 
performs \emph{throughout} the learning process.
We will employ learning algorithms with theoretical 
bounds on the regret for scoring functions that are linear in their parameters, 
making only minimal assumptions about the difference in score between $s^*(x,\bar{y})$ 
and $s^*(x,y)$ in Step 2 of the learning process.
\\~
\textbf{Expert Vs Non-expert user.} We refer to an expert user as
someone who can
demonstrate the optimal trajectory $y^*$ to the robot. For example, robotics
experts such as, the pilot demonstrating helicopter maneuver in Abbeel et
al.~\citep{Abbeel10}. On the other hand, our
non-expert users never demonstrate $y^*$. They can only provide feedback $\bar{y}$
indicating  $s^*(x,\bar{y}) > s^*(x,y)$. For example, users working with
assistive robots on assembly lines.


\label{sec:algorithm}
\section{Learning Algorithm}

For each task, we model the user's scoring function $s^*(x,y)$ with
the following parametrized family of functions. 
\begin{equation}
s(x,y;w) = w \cdot \phi(x,y)
\end{equation}
$w$ is a weight vector that needs to be learned, and $\phi(\cdot)$ are features
describing trajectory $y$ for context $x$. Such linear representation of score functions have been
previously used for generating desired robot
behaviors~\citep{Abbeel10,Ratliff06,Ziebart08}. 

We further decompose the score function in
two parts, one only concerned with the objects the trajectory is interacting with, and the other with the object being 
manipulated and the environment
\begin{align}
\label{eq:scorefn}
\nonumber s(x,y;w_{O},w_{E}) &= s_O(x,y;w_O) + s_E(x,y;w_E)\\
&= w_O \cdot \phi_O(x,y) + w_E \cdot \phi_E(x,y)
\end{align}


We now describe the features for the two terms, $\phi_O(\cdot)$ and $\phi_E(\cdot)$ in the following.

\subsection{Features Describing Object-Object Interactions}
\label{subsec:features}
This feature captures the interaction between
objects in the environment with the object being manipulated. 
We enumerate waypoints of trajectory $y$ as $y_1,..,y_N$ and
objects in the environment as $\mathcal{O}=\{o_1,..,o_K\}$. The robot manipulates the
object $\bar{o}\in \mathcal{O}$. 
A few of the trajectory waypoints would be affected by the other objects in the environment.
For example in Figure~\ref{fig:trajgraph}, $o_1$ and $o_2$ affect 
the waypoint $y_3$ because of proximity.
Specifically, we connect an object $o_k$ to a trajectory waypoint if
the minimum distance to collision is less than a threshold or if $o_k$
lies below $\bar{o}$.  The edge connecting $y_j$ and
$o_k$ is denoted as $(y_j,o_k)\in\mathcal{E}$. 

Since it is the attributes \cite{Koppula11} of the object that really matter in 
determining the trajectory quality, we represent each object with its \emph{attributes}.
Specifically, for every object $o_k$, we consider a vector of $M$ binary
variables [$l_k^1,..,l_k^M$], with each $l_k^m = \{0,1\}$ indicating whether
object $o_k$ possesses property $m$ or not.
  For example, if the set of possible properties are 
\{heavy, fragile, sharp, hot, liquid, electronic\}, then a laptop and a glass table can have 
labels [$0,1,0,0,0,1$] and [$0,1,0,0,0,0$] respectively. 
The
binary variables $l_k^p$ and $l^q$ indicates whether $o_k$ and $\bar{o}$ possess property $p$ and $q$ 
respectively.\footnote{In this work, our goal is to relax the assumption of unbiased 
and close to optimal feedback. We therefore assume complete knowledge of the environment for our algorithm, 
and for the algorithms we compare against. In practice, such knowledge can be extracted using an object attribute
labeling algorithms such as in \cite{Koppula11}.} Then, for every ($y_j,o_k$) edge, we extract following four features 
$\phi_{oo}(y_j, o_k)$: projection of minimum distance to collision along x, y and z (vertical) axis and a binary
variable, that is 1, if $o_k$ lies vertically below $\bar{o}$, 
0 otherwise. 

We now define the score $s_O(\cdot)$ over this graph as follows:
\begin{align}
s_{O}(x,y;w_{O}) =   \sum_{(y_j,o_k)\in\mathcal{E}} \sum_{p,q=1}^M l_k^p l^q
[w_{pq}\cdot\phi_{oo}(y_j, o_k)]
\end{align}
Here, the weight vector
$w_{pq}$ captures interaction between objects with properties $p$ and $q$. 
We obtain $w_O$ in eq.~\eqref{eq:scorefn} by concatenating vectors $w_{pq}$. 
More formally, if the vector at position $i$
of $w_{O}$ is $w_{uv}$ then the vector corresponding to position $i$ of $\phi_O(x,y)$
will be $\sum_{(y_j,o_k)\in\mathcal{E}} l_k^u l^v
[\phi_{oo}(y_j, o_k)]$. 

\subsection{Trajectory Features}

We now describe features, $\phi_{E}(x,y)$, obtained by performing operations on a set of waypoints. 
They comprise the following three types of the features:

\subsubsection{Robot Arm Configurations}
While a robot can reach the same operational
space configuration for its wrist with different configurations of the arm, not all
of them are preferred~\cite{Zacharias11}. For example, the contorted way of holding
the cup shown in Figure~\ref{fig:trajgraph}
may be fine at that time instant, but would present problems if our goal is to 
perform an activity with it, e.g. doing the pouring activity. Furthermore,
 humans like to anticipate robots move and to gain users' confidence, 
robot should produce predictable and legible robotic motions~\citep{Dragan13b}. 

We compute features capturing robot's arm configuration using the location of its
elbow and wrist, w.r.t.\ to its shoulder, in cylindrical  coordinate system, 
$(r, \theta, z)$. We divide a trajectory into three parts in time and compute 9 
features for each of the parts. These features encode the maximum and minimum 
$r$, $\theta$ and $z$ values for wrist and elbow in that part of the trajectory, 
giving us 6 features. Since at the limits of the manipulator configuration, joint 
locks may happen, therefore we also add 3 features for the location of robot's 
elbow whenever the end-effector attains its maximum
$r$, $\theta$ and $z$ values respectively.
Thus obtaining $\phi_{robot}(\cdot) \in \mathbb{R}^{9}$ (3+3+3=9) features for each one-third part 
and $\phi_{robot}(\cdot) \in \mathbb{R}^{27}$ for the complete trajectory.

\begin{figure}[t]
\centering
\includegraphics[width=0.7\linewidth,natwidth=640,natheight=428]{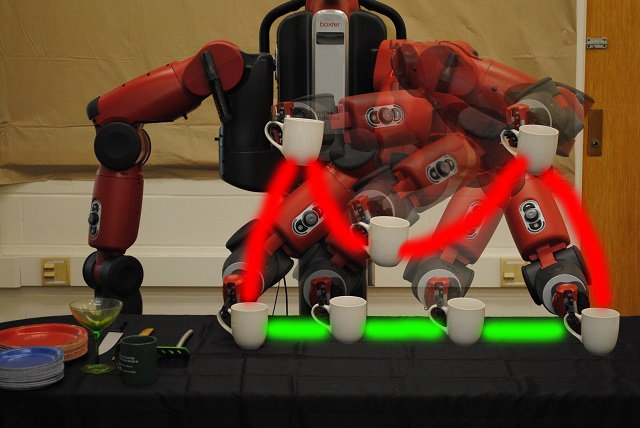}
\includegraphics[width=0.48\linewidth,height=0.7in,clip=true,trim=29 0 39
0,natwidth=242,natheight=388]{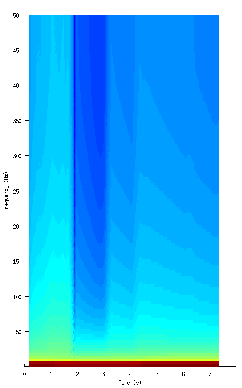}
\includegraphics[width=0.48\linewidth,height=0.7in,clip=true,trim=29 0 30
0,natwidth=242,natheight=388]{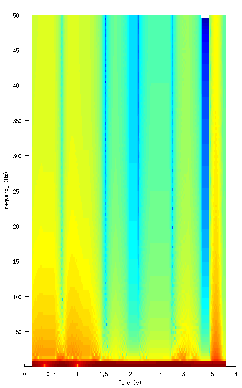}
\caption{\textbf{(Top)} A {\color{green}good} and {\color{red}bad} trajectory for moving a 
mug. The bad trajectory undergoes ups-and-downs. \textbf{(Bottom)} Spectrograms for movement 
in z-direction: \textbf{(Left)} {\color{green}Good} 
trajectory, \textbf{(Right)} {\color{red}Bad} trajectory. }
\label{fig:spect}
\end{figure}

\subsubsection{Orientation and Temporal Behaviour of the Object to be Manipulated}   
{Object orientation during the trajectory is crucial in deciding its quality.} 
For some tasks, the orientation
must be strictly maintained (e.g., moving a cup full of coffee); 
and for some others,
it may be necessary to  change it in a particular fashion (e.g., pouring activity).
Different parts of the trajectory 
may have different requirements over time.
For example, in the placing task, we may
need to bring the object closer to obstacles and be more careful. 

We therefore divide trajectory into three parts in time. 
For each part we store the cosine of the object's maximum 
deviation, along the vertical axis,  from  its final orientation at the goal location. To capture 
object's oscillation along trajectory,
we obtain a spectrogram for each one-third part for the movement of the object
in $x$, $y$, $z$ directions as well as for the deviation along vertical axis (e.g. Figure~\ref{fig:spect}).
We then compute the average power spectral density in the low and high frequency part as  eight 
additional features for each.
This gives us 9 (=1+4*2) features for each one-third part.  Together with one 
additional feature of object's maximum deviation along the whole trajectory,  
we get $\phi_{obj}(\cdot) \in \mathbb{R}^{28}$
(=9*3+1).

\subsubsection{Object-Environment Interactions} 
This feature captures temporal variation of vertical and horizontal distances of the object
$\bar{o}$ from its surrounding surfaces.  In detail, we divide the trajectory into three equal
parts, and for each part we compute object's:
(i) minimum vertical distance from the nearest surface below it. 
(ii) minimum horizontal distance from the surrounding surfaces; and (iii)
minimum distance from the table, on which the task is being performed, and (iv) minimum distance
from the goal location. We also take an average, over all the waypoints, of the horizontal and vertical distances
between the object and the nearest surfaces around it.\footnote{We query PQP collision 
checker plugin of OpenRave for these distances.} 
To capture temporal variation 
of object's distance from its surrounding we plot a time-frequency spectrogram of the object's vertical distance from the nearest surface below it, from which we extract
six features by dividing it into grids. This feature is expressive enough to differentiate
whether an object just grazes over table's edge (steep change in vertical distance) 
versus, it first goes up and over the table and then moves down (relatively smoother change). 
Thus,  the features obtained from object-environment interaction are $\phi_{obj-env}(\cdot) \in 
\mathbb{R}^{20}$ (3*4+2+6=20).

Final feature vector is obtained by concatenating $\phi_{obj-env}$, $\phi_{obj}$ and $\phi_{robot}$,
giving us $\phi_E(\cdot) \in \mathbb{R}^{75}$.

\begin{table*}[t]
\centering
\begin{minipage}{0.28\textwidth}
\flushright
\includegraphics[width=.65\textwidth,natwidth=367,natheight=260]{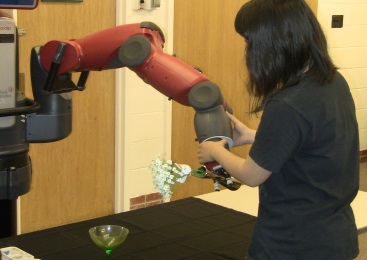}
\includegraphics[width=.65\textwidth,natwidth=360,natheight=215]{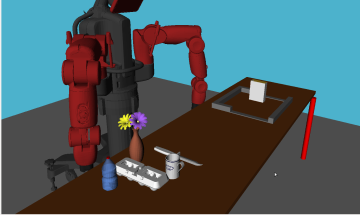}
\end{minipage}
\begin{minipage}{0.1632\textwidth}
\flushright
\includegraphics[width=\textwidth,natwidth=214,natheight=324]{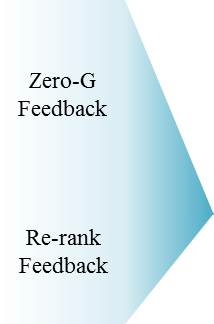}
\end{minipage}
\begin{minipage}{0.45\textwidth}

  \begin{algorithm}[H]
\flushleft
\caption{Trajectory Preference Perceptron. (TPP) } 
\begin{algorithmic}
\STATE Initialize ${w}_O^{(1)} \leftarrow 0$, ${w}_E^{(1)} \leftarrow 0$
\FOR{$t = 1 \;\text{to}\; T$}
	\STATE Sample trajectories $\{y^{(1)},...,y^{(n)}\}$
	\STATE ${y}_t = argmax_{{y}} s(x_t,{y}; w_O^{(t)},w_E^{(t)})$
	\STATE Obtain user feedback $\bar{{y}}_t$
	\STATE $w_O^{(t+1)} \leftarrow w_O^{(t)} + {\phi}_O(x_t,\bar{{y}}_t) -
	{\phi}_O(x_t,{{y}}_t)$
	\STATE $w_E^{(t+1)} \leftarrow w_E^{(t)} + {\phi}_E(x_t,\bar{{y}}_t) - {\phi}_E(x_t,{{y}}_t)$

\ENDFOR
\end{algorithmic}
\label{alg:coactive}
\end{algorithm}
\end{minipage}
\captionof{figure}{Shows our system design, for grocery store settings, which provides users with three choices for iteratively 
improving trajectories. In one type of feedback (\textbf{zero-G} or \textbf{interactive} feedback in case of PR2) user corrects a trajectory 
waypoint directly on the robot while in the second (\textbf{re-rank}) user chooses the top trajectory out of 5 shown on the simulator.}
\label{fig:system-design}
\end{table*}

\subsection{Computing Trajectory Rankings\label{sec:plannerranking}}
For obtaining the top trajectory  (or a top few) for a given task with context $x$, we would like
to maximize the current scoring function $s(x,y;w_O,w_E)$.
\begin{equation}
y^*  = \arg\max_y s(x, {y};w_O,w_E). \label{eq:argmax_rank}
\end{equation}
Second, for a given set $\{y^{(1)},\ldots,y^{(n)}\}$ of 
discrete trajectories, we need to compute ($\ref{eq:argmax_rank}$). 
Fortunately, the latter problem is easy to solve and simply amounts 
to sorting the trajectories by their trajectory scores $s(x,y^{(i)};w_O,w_E)$. 
Two effective ways of solving the former problem are either discretizing the state space~\citep{Alterovitz07,Bhattacharya11,Vernaza12} or directly
sampling trajectories from the continuous space~\citep{Berg10,Dey12}.
Previously, both
approaches have been studied. 
However, for high DoF manipulators the sampling based approach~\citep{Berg10,Dey12}
maintains tractability of the problem, hence we take this 
approach. More precisely, similar to~\citet{Berg10},
we sample trajectories using rapidly-exploring random trees (RRT)~\citep{Lavalle01}.\footnote{When RRT becomes too slow, we switch to a more efficient bidirectional-RRT.The cost function (or its approximation) we learn can be fed to 
trajectory optimizers like CHOMP~\citep{Chomp} or optimal planners like RRT*~\citep{Karaman10} to produce reasonably good trajectories.}
However, naively sampling trajectories could return many similar trajectories. To get diverse 
samples of trajectories we use various diversity introducing methods.
For example, we introduce obstacles in the environment which forces the planner
to sample different trajectories. Our methods also introduce randomness in
planning by initizaling goal-sample bias of RRT planner randomly. To avoid
sampling similar trajectories multiple times, one of our diversity method
introduce obstacles to block waypoints of
already sampled trajectories.  
Recent work by Ross et al.~\citep{Ross13} propose
the use of sub-modularity to achieve diversity. For more details on sampling trajectories we
refer interested readers to the work by Erickson and LaValle~\citep{Erickson09},
and Green and Kelly~\citep{Green11}.
Since our primary goal is to learn a score function on 
trajectories we now describe our learning algorithm.

\subsection{Learning the Scoring Function} 
\label{sec:learningscore}


\noindent The goal is to learn the parameters $w_O$ and $w_E$
of the scoring function $s(x,y;w_O,w_E)$  
so that it can be used to 
rank trajectories according to the user's preferences. To do so, we adapt the Preference 
Perceptron algorithm~\cite{Shivaswamy12} as detailed in Algorithm~\ref{alg:coactive}, and we call it the Trajectory Preference Perceptron (TPP). 
Given a context $x_t$, the top-ranked trajectory $y_t$ under the current parameters $w_O$ and $w_E$, and the user's feedback trajectory $\bar{y}_t$, the TPP updates the weights in the direction 
${\phi}_O(x_t,\bar{y}_t) - {\phi}_O(x_t,{y}_t)$ and
${\phi}_E(x_t,\bar{y}_t) - {\phi}_E(x_t,{y_t})$ respectively. Our
update equation resembles to the weights update equation in Ratliff et
al.~\citep{Ratliff06}. However, our update does not depends on the availability of optimal demonstraions.  Figure~\ref{fig:system-design} shows an overview of our system design.

Despite its simplicity and even though the algorithm typically does not receive the optimal 
trajectory $y_t^* = \arg\max_ys^*(x_t,y)$ as feedback, the \textit{TPP enjoys guarantees on 
the regret}~\cite{Shivaswamy12}.
We merely need to characterize by how much the feedback improves on the presented 
ranking using the following definition of expected $\alpha$-informative feedback:
$$
E_t[s^*(x_t,\bar{y}_t) ] \geq s^*(x_t,y_t) + 
\alpha ( s^*(x_t,y^*_t) - s^*(x_t,y_t)) - \xi_t
$$
This definition states that the user feedback should have a score of ${\bar{y}}_t$ that is --
in expectation over the users choices -- higher than that of ${y}_t$ by a fraction $\alpha
\in (0,1]$ of the maximum possible range $s^*(x_t,{\bar{y}}_t) -
s^*(x_t,{y}_t)$. It is important to note that this condition only
needs to be met
in expectation and not deterministically. This leaves room for noisy and
imperfect user feedback. If
this condition is not fulfilled due to bias in the feedback, the slack variable $\xi_t$ captures the
amount of violation. In this way any feedback can be described by an appropriate combination of 
$\alpha$ and $\xi_t$.
Using these two parameters, 
the proof by Shivaswamy and Joachims~\citep{Shivaswamy12} can be adapted
(for proof see Appendix~\ref{subsec:proof} \& \ref{subsec:proof-cor}) to show
that average regret of TPP is upper bounded by:
$$E[REG_T] \le \mathcal{O} (\frac{1}{\alpha\sqrt{T}} + \frac{1}{\alpha
T}\sum_{t=1}^T \xi_t )$$

In practice, over feedback iterations the quality of trajectory $y$  proposed by
robot improves. The $\alpha$-informative criterion only requires the user to  improve
$y$ to $\bar{y}$ in expectation. 



\section{Experiments and Results}
\label{sec:experiment}
We first describe our experimental setup, then 
present quantitative results (Section~\ref{sec:results}) ,
and then present robotic experiments on PR2 and Baxter (Section~\ref{sec:user-study}). 

\subsection{Experimental Setup}

\header{Task and Activity Set for Evaluation.}
We evaluate our approach on 35 robotic tasks in a household setting and 16 pick-and-place tasks in a grocery store checkout 
setting. For household activities we use PR2, and for the grocery store setting
we use Baxter. To assess the generalizability of our approach, for each task we train and test on scenarios 
with different objects being manipulated and/or with a different environment.
We evaluate the quality of trajectories after the robot has grasped the item in
question and 
while the robot moves it for task completion. Our work complements previous works on
	grasping items \citep{Saxena08,Lenz13}, 
pick and place tasks~\citep{Jiang12b}, and
detecting bar codes for grocery checkout~\citep{Klingbeil11}.
We consider the following three most commonly occurring activities in household and grocery stores:

\begin{figure*}[t]

\begin{tabular}{ccc}
\cellcolor{red}\textit{Manipulation centric}&\cellcolor{green}\textit{Environment centric}&\cellcolor{blue}\textit{Human centric}\\
\subfigure[\hspace*{-1mm}\scriptsize{Moving flower vase}]{
\includegraphics[width=0.3\textwidth,height=0.22\textwidth,natwidth=945,natheight=645]{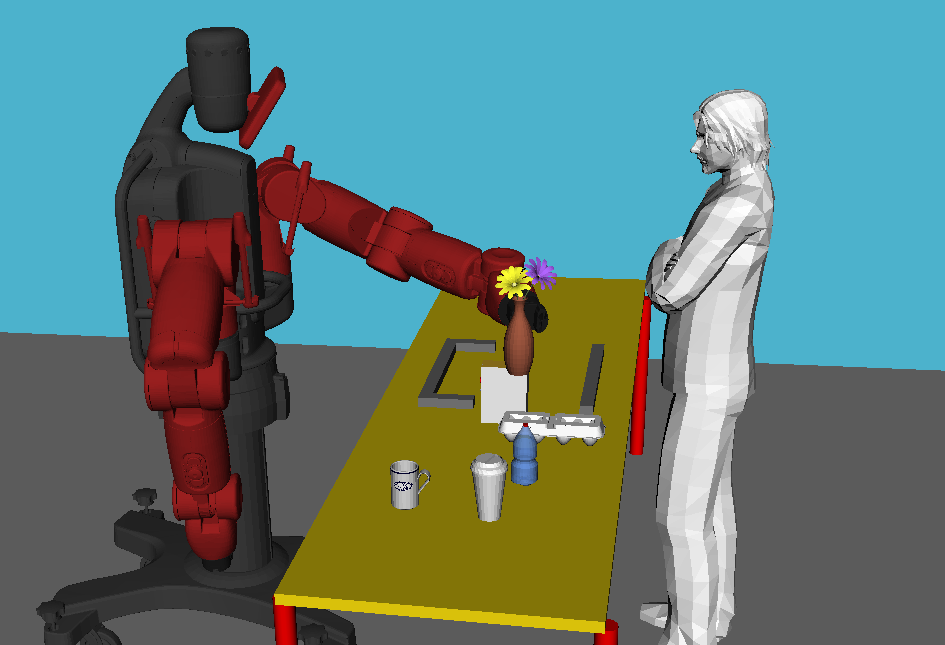}
}
&
\subfigure[\hspace*{-1mm}\scriptsize{Checking out eggs}]{
\includegraphics[width=0.3\textwidth,height=0.22\textwidth,natwidth=829,natheight=587]{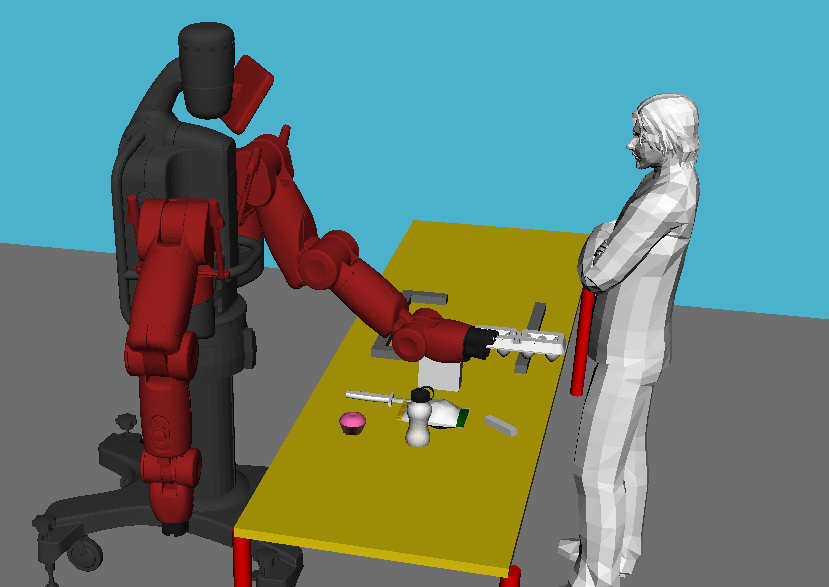}
}
&
\subfigure[\hspace*{-1mm}\scriptsize{Manipulating knife}]{
\includegraphics[width=0.3\textwidth,height=0.22\textwidth,natwidth=1114,natheight=666]{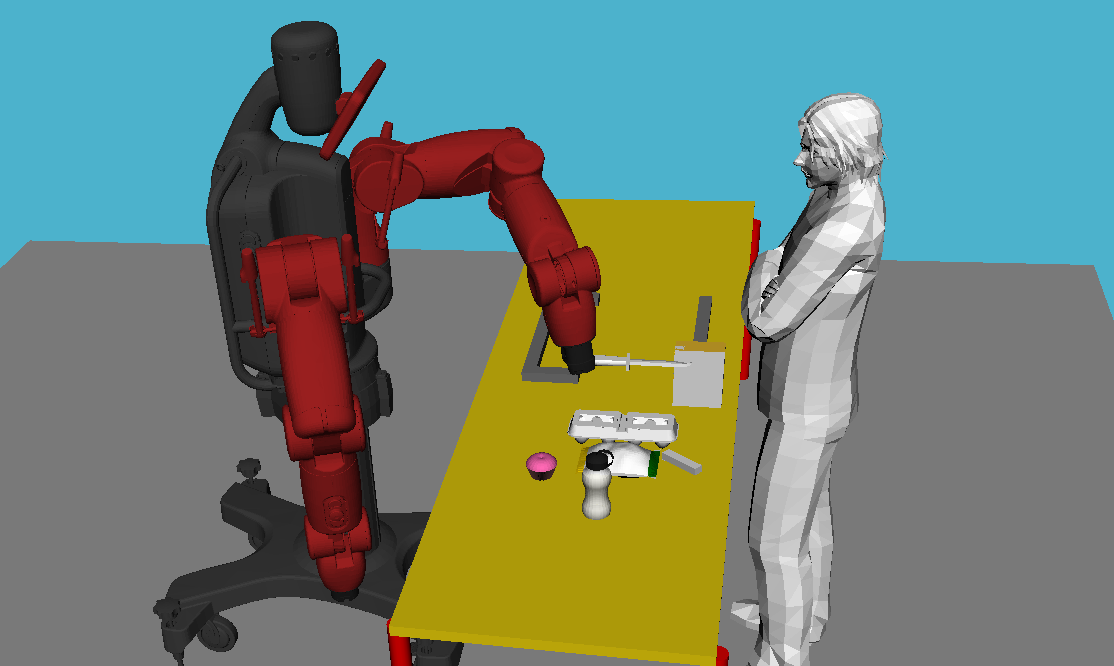}
}\\
\multicolumn{3}{c}{Baxter in a grocery store setting.}
\\
\setcounter{subfigure}{0}
\subfigure[\hspace*{-1mm}\scriptsize{Pouring water}]{
\includegraphics[width=0.3\textwidth,height=0.24\textwidth,natwidth=843,natheight=741]{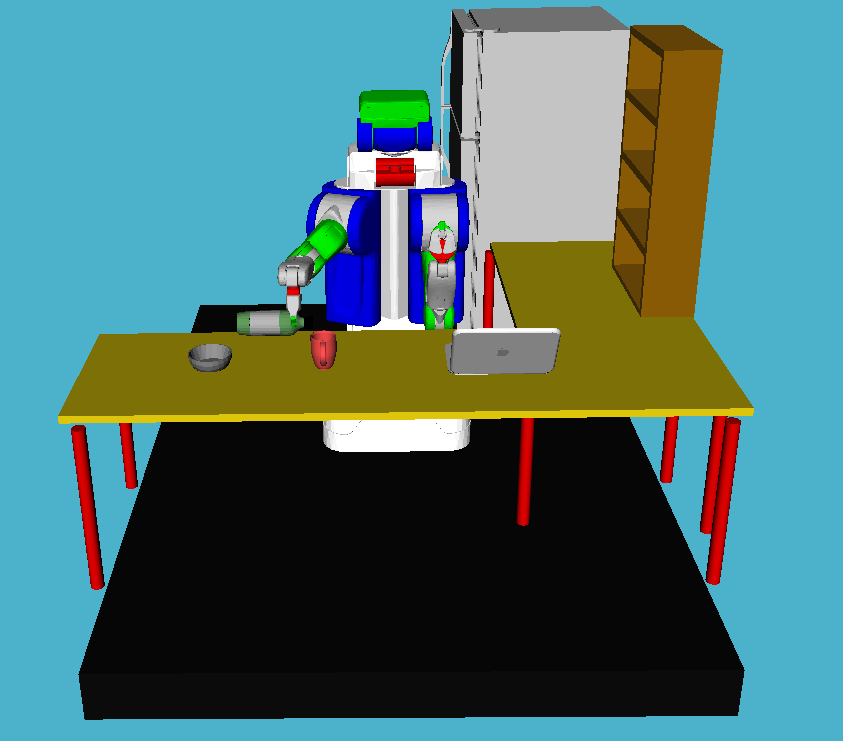}
}
&
\subfigure[\hspace*{-1mm}\scriptsize{Moving liquid near laptop}]{
\includegraphics[width=0.3\textwidth,height=0.24\textwidth,natwidth=1225,natheight=965]{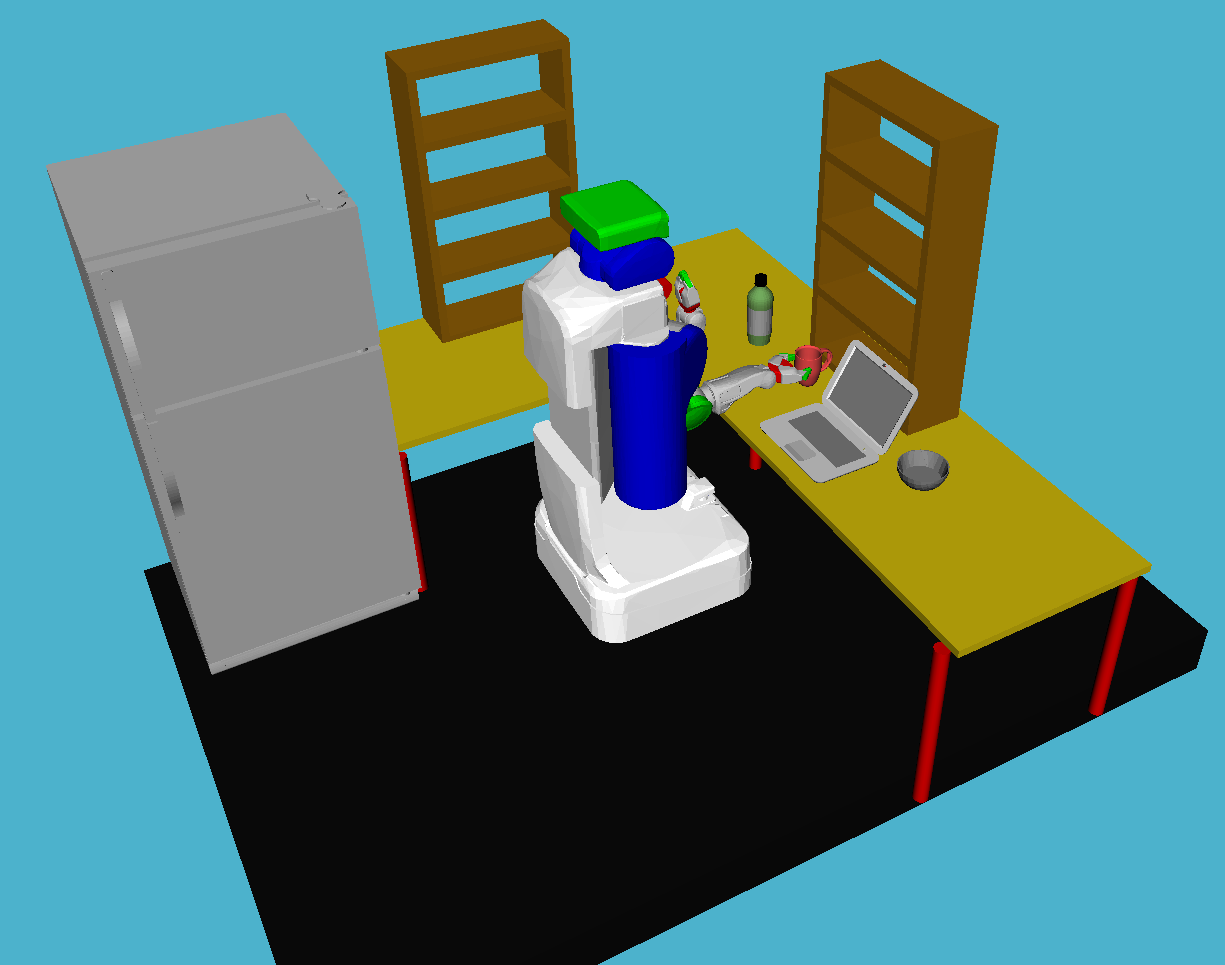}
}
&
\subfigure[\hspace*{-1mm}\scriptsize{Manipulating sharp object}]{
\includegraphics[width=0.3\textwidth,height=0.24\textwidth,natwidth=1072,natheight=782]{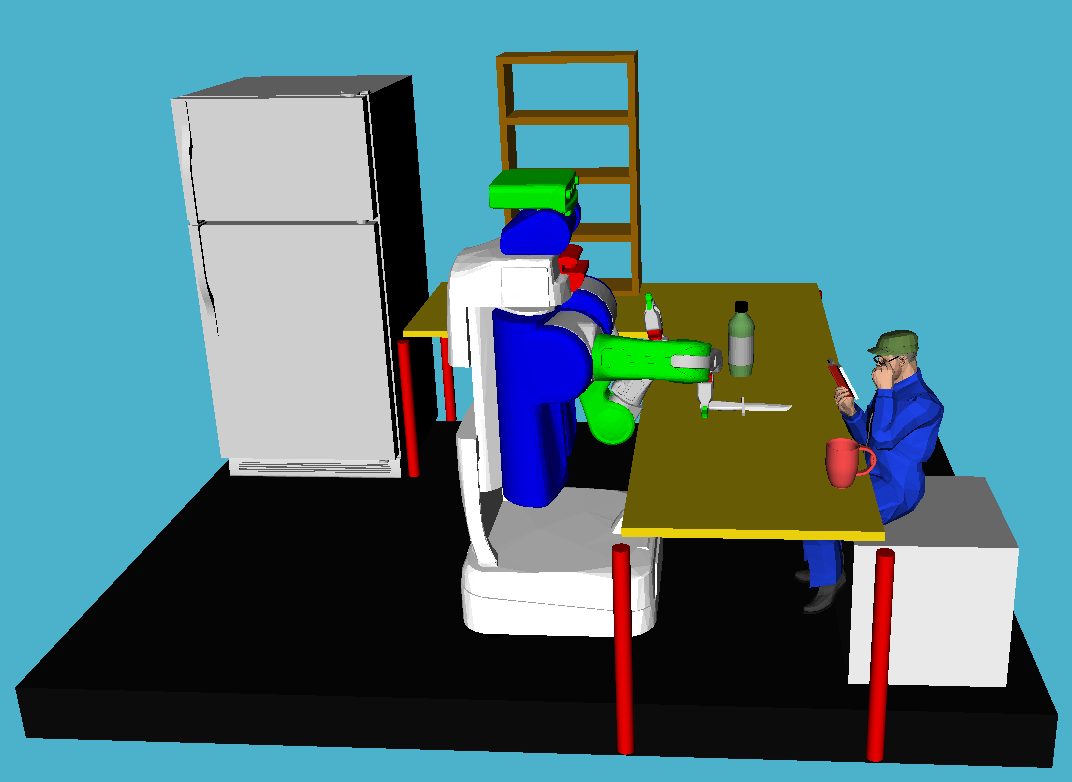}
}\\
\multicolumn{3}{c}{PR2 in a household setting.}
\end{tabular}
\caption{Robot demonstrating different grocery store and household activities with various 
objects (\textbf{Left}) \textit{Manipulation centric:} while pouring water the tilt angle of bottle 
must change in a particular manner, similarly a flower vase should be kept upright. (\textbf{Middle}) 
\textit{Environment centric:} laptop is an electronic device so robot must carefully move water near it, similarly eggs are
fragile and should not be lifted too high. 
(\textbf{Right}) \textit{Human centric:} knife is sharp and 
interacts with nearby soft items and humans. It should strictly be kept at a safe distance from humans. (\textbf{Best viewed in color})}
\label{fig:taskcats}
\end{figure*}

\begin{enumerate}
\item \emph{Manipulation centric:} These activities are primarily concerned with the object 
being manipulated. 
Hence the object's properties and the way the robot moves it in the environment 
are more relevant. Examples of such household activities are pouring water into
a cup or inserting pen into a pen holder, as in Figure~\ref{fig:taskcats} (Left). 
While in a grocery store, such activities could include moving a flower vase or
moving fruits and vegetables, which could be damaged
 when dropped or pushed into other items. We consider \textit{pick-and-place,
pouring and inserting activities} with following objects: cup, bowl, bottle,
pen, cereal box, flower vase, and tomato. Further, in every environment we place many objects, along with the object to be manipulated, to restrict simple straight line trajectories. 
\item \emph{Environment centric:} These activities are also concerned with the interactions of the object
being manipulated with the surrounding objects. Our object-object interaction
features (Section~\ref{subsec:features}) allow the algorithm to learn preferences on trajectories for moving
fragile objects like egg cartons or moving liquid near electronic devices, as in Figure~\ref{fig:taskcats} (Middle).
We consider moving fragile items like egg carton, heavy metal boxes near a glass table, water near laptop and other electronic devices.
\item \emph{Human centric:} Sudden movements by the robot put the human in danger of getting hurt. We consider activities 
where a robot manipulates sharp objects such as knife, as in Figure~\ref{fig:taskcats}
(Right), moves a hot coffee cup or a bowl of water with a human in vicinity.
\end{enumerate}

\header{Experiment setting.} Through experiments we will study:
\begin{itemize}
\item \textit{Generalization}: Performance of robot on tasks that it has not
seen before.
\item \textit{No demonstrations}:  Comparison of TPP to algorithms that
also learn in absence of expert's demonstrations.
\item \textit{Feedback:} Effectiveness of different kinds of user feedback in
absence of expert's demonstrations.
\end{itemize}

\begin{figure*}[t]
\centering
\begin{tabular}{@{}l@{}c@{}c@{}c@{}}
&
\scriptsize{Same environment, different object.}
&
\scriptsize{New Environment, same object.}
&
\scriptsize{New Environment, different object.}\\
\raisebox{.5in}{\rotatebox[origin=c]{90}{\scriptsize nDCG@3}}
&
\subfigure{
\includegraphics[width=0.32\linewidth,height=0.22\textwidth,natwidth=941,natheight=720]{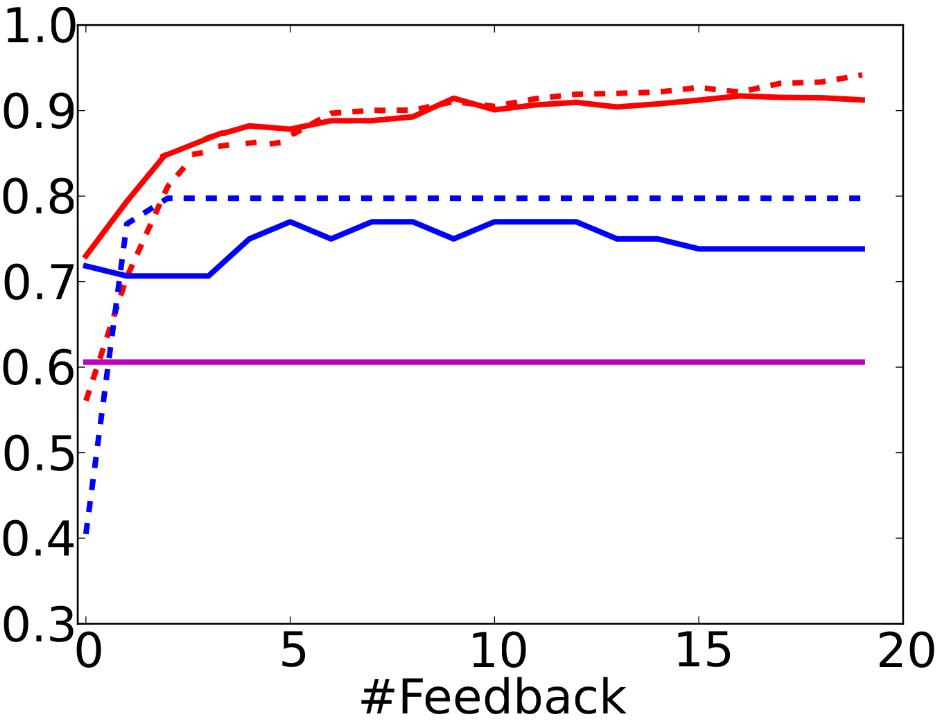}
\label{fig:genobj}
}
&
\subfigure{
\includegraphics[width=0.32\linewidth,height=0.22\textwidth,natwidth=937,natheight=722]{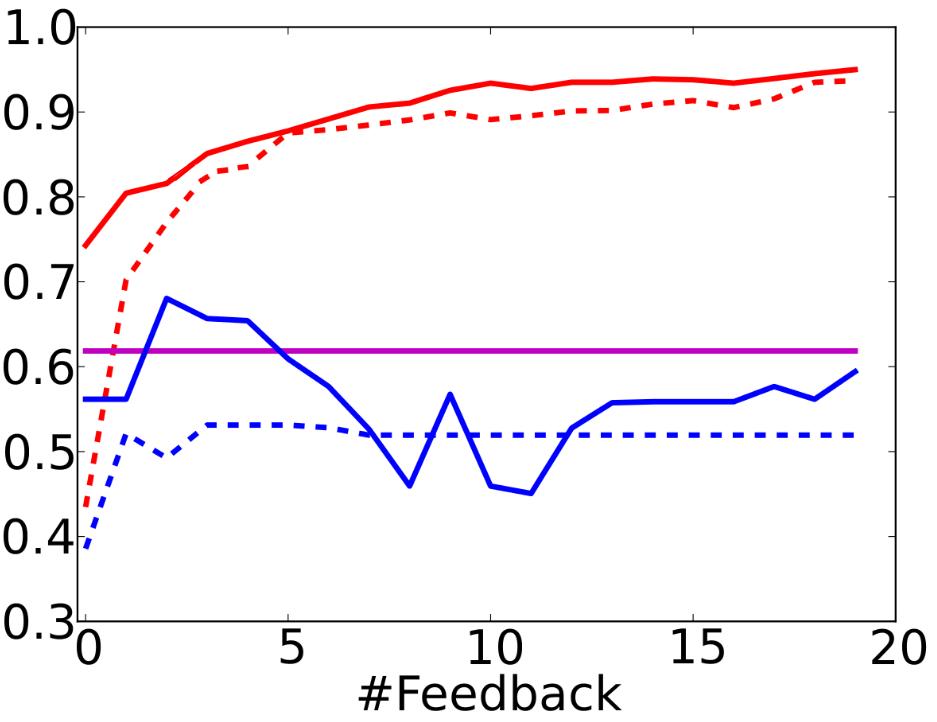}
\label{fig:genenv}
}
&
\subfigure{
\includegraphics[width=0.32\linewidth,height=0.22\textwidth,natwidth=942,natheight=720]{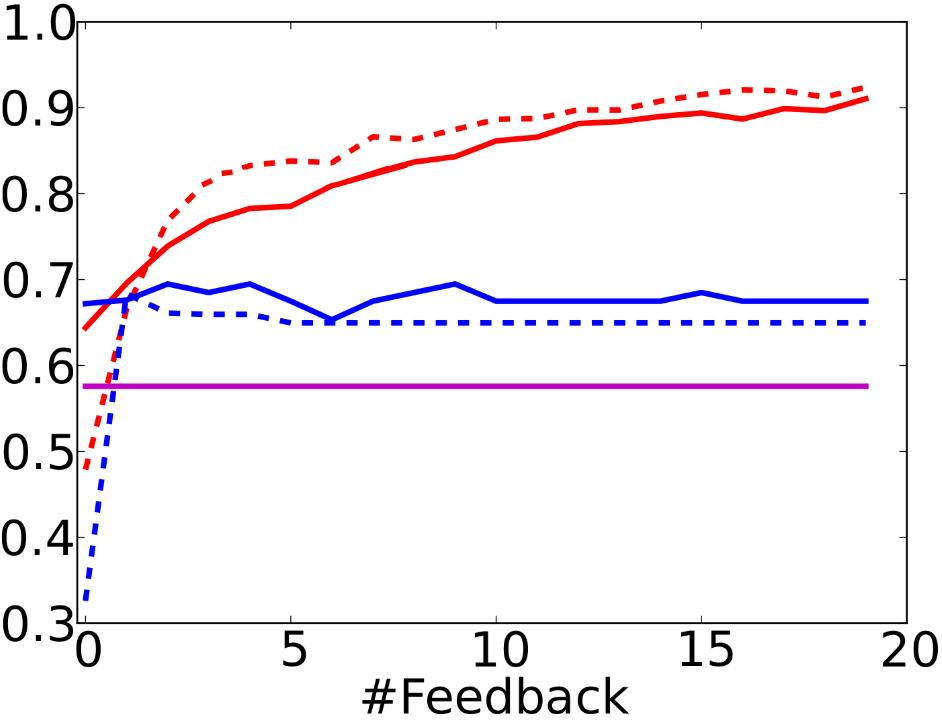}
\label{fig:genenvobj}
}
\end{tabular}
\caption*{Results on Baxter in grocery store setting.}
\begin{tabular}{@{}l@{}c@{}c@{}c@{}}
\raisebox{.5in}{\rotatebox[origin=c]{90}{\scriptsize nDCG@3}}
&
\subfigure{
\includegraphics[width=0.32\linewidth,height=0.22\textwidth,natwidth=897,natheight=718]{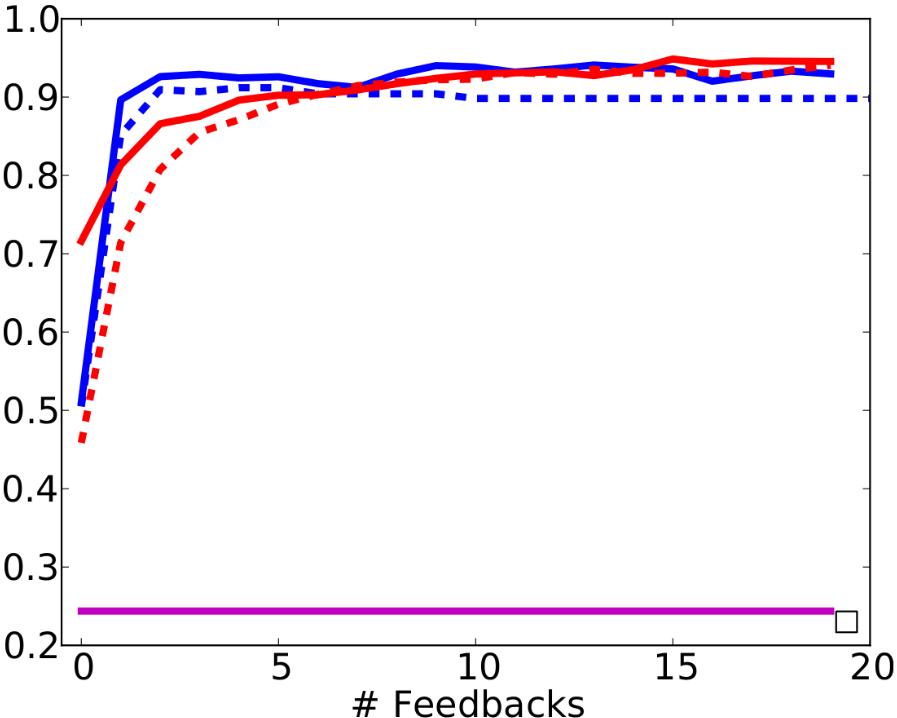}
\label{fig:genobj}
}
&
\subfigure{
\includegraphics[width=0.32\linewidth,height=0.22\textwidth,natwidth=893,natheight=718]{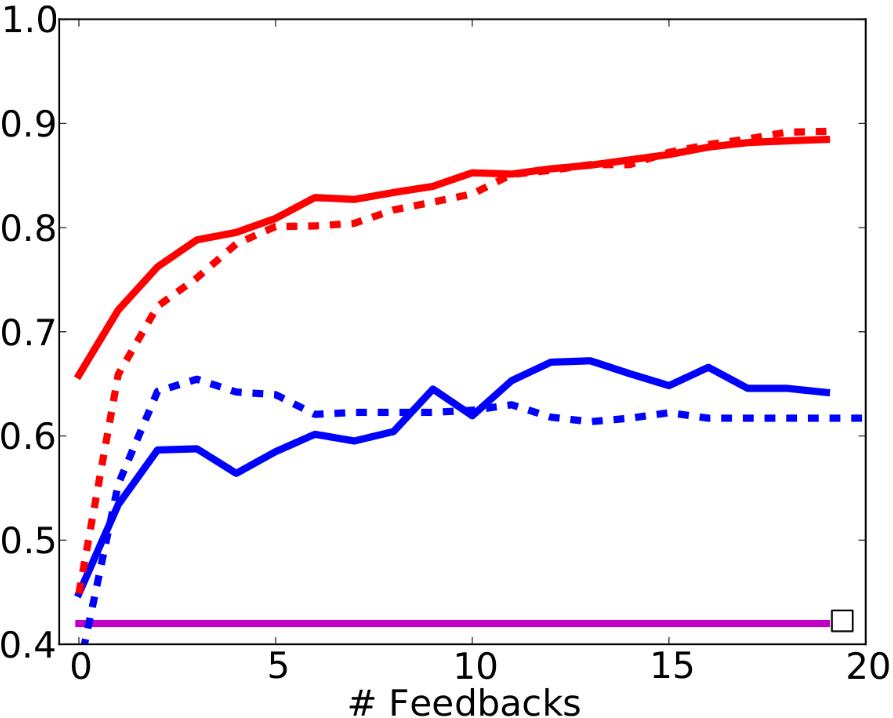}
\label{fig:genenv}
}
&
\subfigure{
\includegraphics[width=0.32\linewidth,height=0.22\textwidth,natwidth=893,natheight=718]{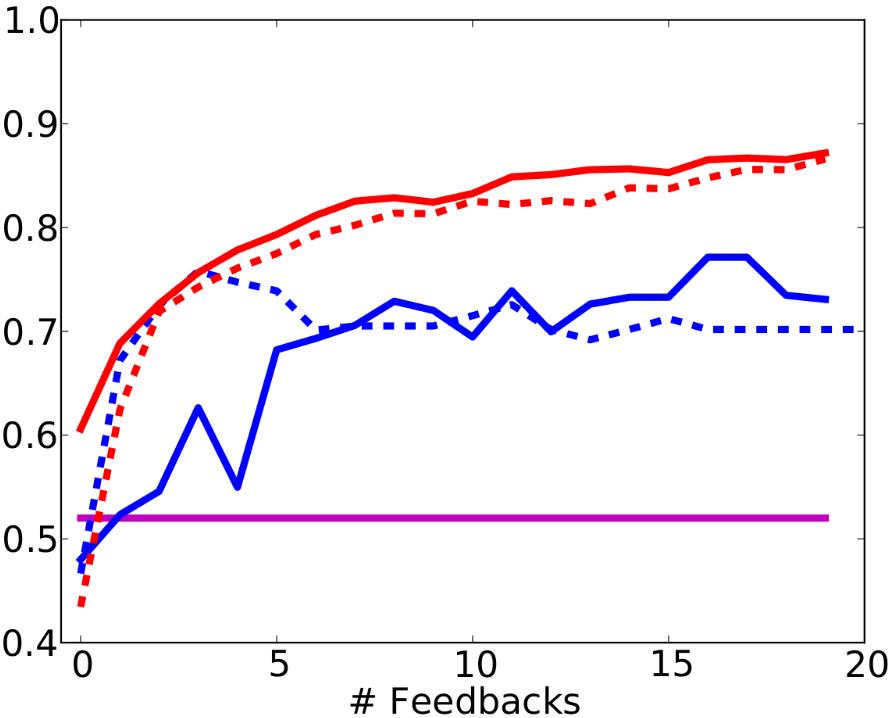}
\label{fig:genenvobj}
}
\end{tabular}
\caption*{Results on PR2 in household setting.}
\caption{{Study of generalization with change in object, environment and both.} {\color{magenta}{Manual}}, \textcolor{blue}{Pre-trained MMP-online (---)}, \textcolor{blue}{Untrained MMP-online (-- --)}, \textcolor{red}{ Pre-trained TPP (---)}, \textcolor{red}{ Untrained TPP (-- --)}. }
\label{fig:genstudy}
\end{figure*}
\header{Baseline algorithms.} 
We evaluate algorithms that learn preferences from online feedback under two settings:
(a) \emph{untrained}, where the algorithms learn preferences 
for a new task from scratch without observing any previous feedback; 
(b) \emph{pre-trained}, where the algorithms are pre-trained 
on other similar tasks, and then adapt 
to a new task.
We compare the following algorithms:

\begin{itemize}
\item \emph{Geometric}: The robot plans a path, independent of the task, using a
Bi-directional RRT (BiRRT)~\citep{Lavalle01} planner.
\item \emph{Manual}:  The robot plans a path following certain manually coded preferences.
\item \emph{TPP}:  Our algorithm, evaluated under both \emph{untrained} and \emph{pre-trained} settings.
\item \emph{MMP-online}: This is an online implementation of the Maximum Margin
Planning (MMP)~\citep{Ratliff06,Ratliff09a}
algorithm. MMP attempts to make an expert's trajectory better than any other trajectory by a 
margin. It can be interpreted as a special case of our algorithm with
1-informative i.e. optimal feedback. However, directly adapting
MMP~\citep{Ratliff06} to our experiments poses two
challenges: (i) we do not have knowledge of the optimal trajectory; and (ii) the state space 
of the manipulator we consider is too large, discretizing which makes 
intractable to train MMP.

 To ensure a fair comparison, we follow the MMP
algorithm from~\citep{Ratliff06,Ratliff09a} and train it under similar settings
as TPP. Algorithm~\ref{alg:mmp} shows
our implementation of MMP-online. It is very similar to TPP
(Algorithm~\ref{alg:coactive}) but with a different parameter update step. 
Since both algorithms only observe user feedback and not demonstrations,
MMP-online treats each feedback as a proxy for optimal demonstration.
At every iteration MMP-online trains a structural support vector machine
(SSVM)~\cite{Joachims09} using all previous feedback as training examples, 
and use the learned weights to predict trajectory scores in the next iteration. 
Since the argmax operation is performed on a set of trajectories it
remains tractable. We quantify closeness of trajectories by the $L_2-$norm
of the difference in 
their feature representations, and choose the regularization
parameter $C$ for training SSVM in hindsight, giving an unfair advantage to MMP-online.

\end{itemize}

\begin{algorithm}
\caption{MMP-online} 
\begin{algorithmic}

\STATE Initialize ${w}_O^{(1)} \leftarrow 0$, ${w}_E^{(1)} \leftarrow 0,
\mathcal{T}=\{\}$
\FOR{$t = 1 \;\text{to}\; T$}
	\STATE Sample trajectories $\{y^{(1)},...,y^{(n)}\}$
	\STATE ${y}_t = argmax_{{y}} s(x_t,{y}; w_O^{(t)},w_E^{(t)})$
	\STATE Obtain user feedback $\bar{{y}}_t$
    \STATE $\mathcal{T} = \mathcal{T}\cup \{(x_t,\bar{y}_t)\}$
	\STATE $w_O^{(t+1)}, w_E^{(t+1)} =\text{Train-SSVM}(\mathcal{T})$\;(Joachims et al.~\citep{Joachims09})
\ENDFOR

\end{algorithmic}
\label{alg:mmp}
\end{algorithm}
\header{Evaluation metrics.}
In addition to performing a user study (Section~\ref{sec:user-study}), we also 
designed two datasets to quantitatively evaluate the performance of our online algorithm. 
We obtained experts labels on 1300 trajectories in a grocery setting and 2100
trajectories in a household setting.
Labels were on the basis of subjective 
human preferences on a Likert scale of 1-5 (where 5 is the best). 
Note that these absolute ratings are never provided
to our algorithms and are only used for the quantitative evaluation of different algorithms.

We evaluate performance of algorithms by measuring how well they rank
trajectories,
that is, trajectories with higher Likert score should be ranked higher.
To quantify the quality of a ranked list of trajectories we report normalized
discounted cumulative gain (nDCG)~\citep{Manning08} ---
criterion popularly used in Information Retrieval for document ranking.
In particular we report nDCG at positions 1 and 3, equation~\eqref{eq:ndcg}.
While nDCG@1 is a suitable metric for autonomous robots that execute the top
ranked
trajectory (e.g., grocery checkout), nDCG@3 is suitable for scenarios where the
robot is supervised
by humans, (e.g., assembly lines). For a given ranked list of items
(trajectories here) nDCG at position $k$ is defined as:
\begin{align}
DCG@k &= \sum_{i=1}^{k}\frac{l_i}{\log_2(i+1)}\\
\label{eq:ndcg} nDCG@k &= \frac{DCG@k}{IDCG@k},
\end{align}
where $l_i$ is the Likert score of the item at position $i$ in
the ranked list. $IDCG$ is the $DCG$ value of the best possible ranking of
items.
It is obtained by ranking items in decreasing order of their Likert score.

\subsection{Results and Discussion}
\label{sec:results}

\noindent We now present quantitative results where we compare TPP against the
baseline algorithms on our data set of labeled trajectories.  
\\~\\~
\header{How well does TPP generalize to new tasks?} To study generalization of preference feedback
we evaluate performance of TPP-pre-trained (i.e., \textit{TPP} algorithm under \textit{pre-trained} setting)
 on a set of tasks the algorithm has not seen before.
We study generalization when: (a) only the object being
manipulated changes, e.g., a bowl replaced by a cup or an egg carton replaced by tomatoes, (b)
only the surrounding environment changes, e.g., rearranging objects in the
environment or changing the start location of tasks, and (c) when both change.
Figure~\ref{fig:genstudy} shows nDCG@3 plots averaged over tasks for all types of activities for both household and grocery store settings.\footnote{Similar results were obtained with nDCG@1 metric.}
TPP-pre-trained starts-off with higher nDCG@3 values than TPP-untrained in all three cases. 
However, as more feedback is provided, the performance of both
algorithms improves, and they eventually give identical performance. 
We further observe that generalizing to tasks with both new environment and object is harder than when only 
one of them changes.
\\~\\~
\header{How does TPP compare to MMP-online?} MMP-online while training assumes
all user feedback is
optimal, and hence over time it accumulates many contradictory/sub-optimal training examples. 
We empirically observe that MMP-online generalizes better in grocery store setting than the household setting (Figure~\ref{fig:genstudy}), however under
both settings its performance remains much lower than TPP. 
This also highlights the sensitivity of MMP to sub-optimal demonstrations.  
\\~\\~
\header{How does TPP compare to Manual?} For the manual baseline we encode some
preferences into the planners, e.g., keep 
a glass of water upright. However, some preferences are difficult to specify, e.g., not to move heavy objects over fragile 
items. We empirically found (Figure~\ref{fig:genstudy}) that the resultant manual algorithm produces poor trajectories in
comparison with TPP, with
an average nDCG@3 of 0.44 over all types of household activities. 
Table~\ref{tab:algostudy} reports nDCG values averaged over 20 feedback
iterations in untrained setting. For both household and 
grocery activities, TPP performs better than other baseline algorithms. 
\\~\\~
\begin{figure}[t]
\centering
\includegraphics[width=.8\linewidth,natwidth=1202,natheight=925]{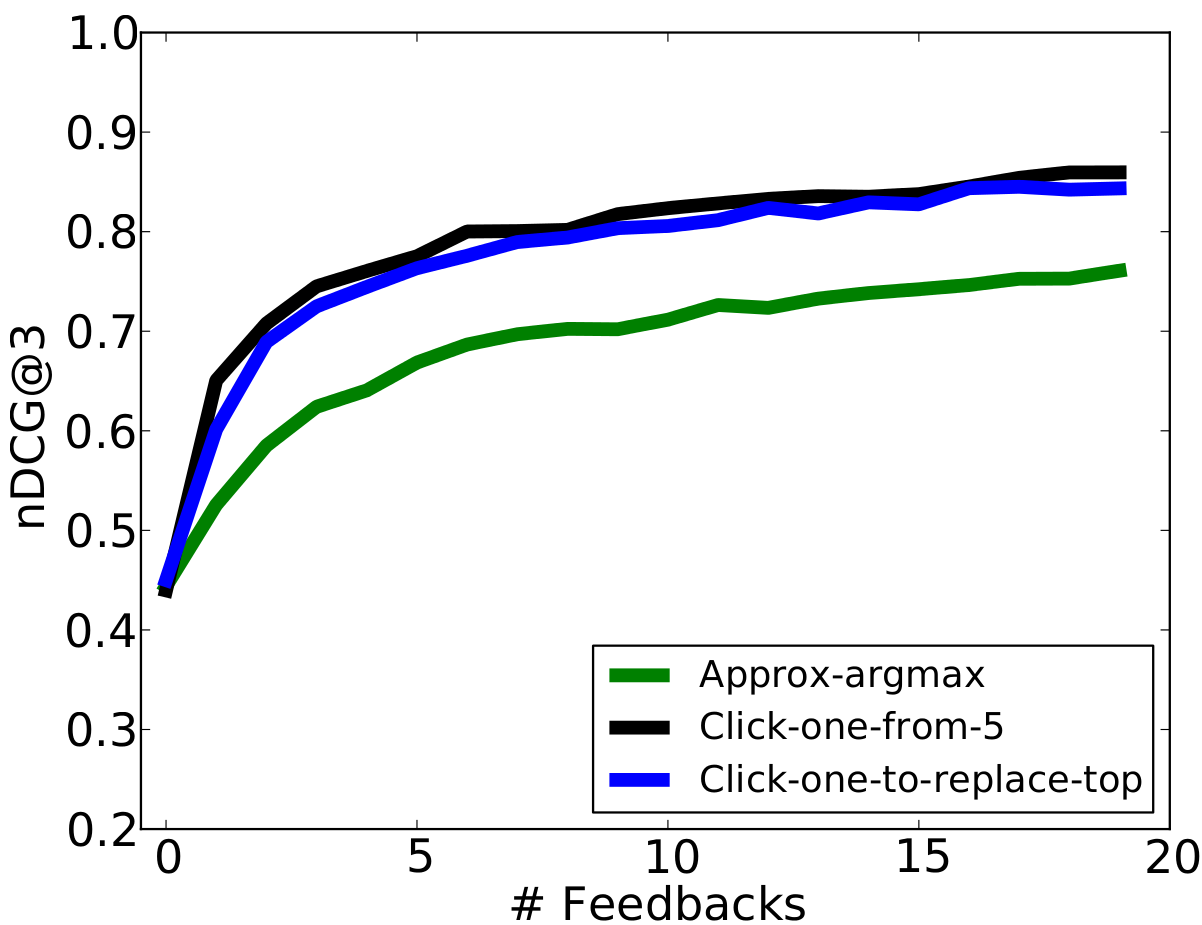}
\caption{{Study of re-rank feedback on Baxter for grocery store setting.}}
\label{fig:feedback}
\end{figure}
\header{How does TPP perform with weaker feedback?} To study  
the robustness of TPP to less informative feedback we consider the following variants of re-rank feedback:
\begin{itemize}
\item \textit{Click-one-to-replace-top:} User observes the trajectories sequentially in order of their current 
predicted scores and clicks on the first trajectory which is better than the top ranked 
trajectory. 
\item \textit{Click-one-from-5:}  Top $5$ trajectories are shown and user clicks on the one he thinks
is the best after watching all $5$ of them.
\item \textit{Approximate-argmax}: This is a weaker feedback, here instead of presenting top ranked 
trajectories, five  random trajectories are selected 
as candidate. The user selects the best trajectory among these 5 candidates.
This simulates a situation when computing an argmax over trajectories is prohibitive and therefore 
approximated.
\end{itemize}
\begin{table*}[t]
\centering
 \begin{tabular}{r@{}rr|c@{\hskip 0.1in}c@{\hskip 0.1in}c@{\hskip 0.1in}c|c@{\hskip 0.1in}c@{\hskip 0.1in}c@{\hskip 0.1in}c}
 \cline{2-11}
 &&&\multicolumn{4}{c|}{Grocery store setting on Baxter.} & \multicolumn{4}{c}{Household setting on PR2.} \\
 \cline{4-11}
 &&\multirow{2}{*}{Algorithms}&Manip. &Environ.&Human&\multirow{2}{*}{Mean}&Manip. &Environ.&Human&\multirow{2}{*}{Mean}\\
 &&&centric&centric&centric&&centric&centric&centric&\\
 \cline{2-11}
 &&Geometric&.46  (.48)&.45  (.39)&.31  (.30)&.40  (.39)  &.36 (.54)&.43 (.38)&.36 (.27)&.38 (.40)\\
 &&Manual&.61 (.62)&.77  (.77)&.33  (.31)&.57  (.57) &.53 (.55)&.39 (.53)&.40 (.37)&.44 (.48)  \\
 &&MMP-online&.47 (.50)&.54 (.56)&.33 (.30)&.45 (.46) &.83 (.82)&.42 (.51)&.36 (.33)&.54 (.55)  \\
 &&TPP&\textbf{.88 (.84)}&\textbf{.90 (.85)}&\textbf{.90 (.80)}&\textbf{.89 (.83)} &\textbf{.93 (.92)}&\textbf{.85 (.75)}&\textbf{.78 (.66)}&\textbf{.85 (.78)}\\
 \cline{2-11}
 \end{tabular}

\caption{\textbf{Comparison of different algorithms in untrained setting.} 
Table contains nDCG@1(nDCG@3) values averaged over 20 feedbacks.}
 \label{tab:algostudy}

\end{table*}
Figure~\ref{fig:feedback} shows the performance of TPP-untrained receiving different
kinds of feedback and averaged over three types of activities in grocery store setting. When feedback is more 
$\alpha$-informative the algorithm requires fewer iterations to learn preferences. In particular, 
click-one-to-replace-top and click-one-from-5 
are more informative than approximate-argmax 
and therefore require less feedback to reach a given nDCG@1 value. 
Approximate-argmax improves slowly since it is least informative.
In all three cases the feedback is $\alpha$-informative, for
some $\alpha > 0$, therefore TPP-untrained eventually learns the user's preferences. 

\subsection{Comparison with fully-supervised algorithms} 
\label{subsec:oracle}

The algorithms discussed so far only observes ordinal feedback where the users
iteratively improves upon the proposed trajectory. In this section we compare TPP to a fully-supervised algorithm that observes
expert's labels while training. 
Eliciting such expert labels on the large space of
trajectories is not realizable in practice. However, empirically it
nonetheless provides an upper-bound on the generalization to new tasks.
We refer to this algorithm as
\textit{Oracle-svm} and it learns to rank trajectories using
SVM-rank~\citep{Joachims06}. Since expert labels are not available while
prediction, on test set Oracle-svm
predicts once and does not learn from user feedback. 

\begin{figure}[t]
\centering
\includegraphics[width=.8\linewidth,natwidth=1193,natheight=920]{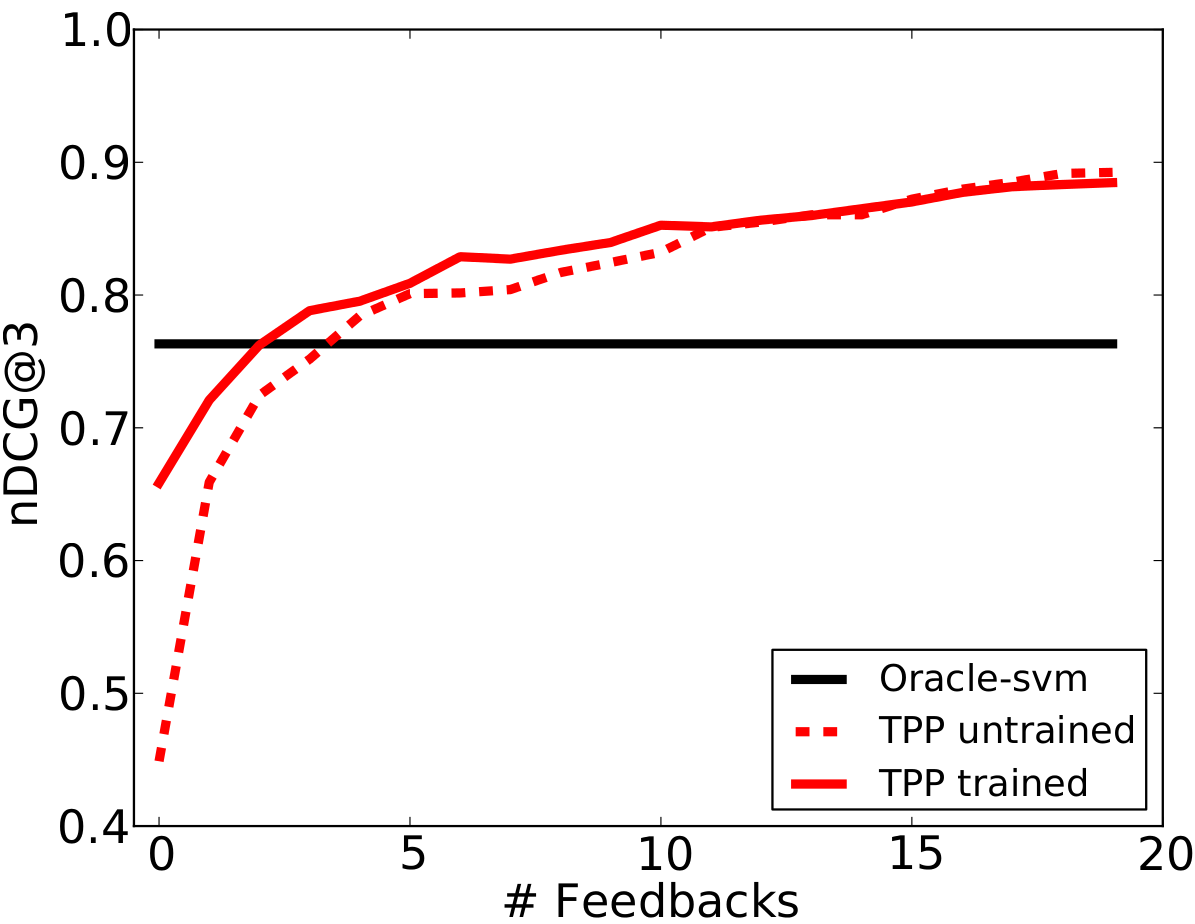}
\caption{{Comparision with fully-supervised Oracle-svm on Baxter for
grocery store setting.}}
\label{fig:oracle-svm}
\end{figure}

Figure~\ref{fig:oracle-svm}
 compares TPP and Oracle-svm on new tasks. Without observing any feedback on new
tasks Oracle-svm performs better than TPP. However, after few feedback iterations
TPP improves over Oracle-svm, which is not updated since it requires expert's labels on test set.
On average, we observe, it takes 5 feedback iterations for TPP to
improve over Oracle-svm. Furthermore, learning from demonstration (LfD) can be seen as a special case of Oracle-svm where,
instead of providing an expert label for every sampled trajectory, the expert
directly demonstrates the optimal trajectory.

\subsection{Robotic Experiment: User Study in learning trajectories}
\label{sec:user-study}
\begin{figure*}[t]
\centering
\includegraphics[width=0.35\textwidth,height=0.26\textwidth,natwidth=576,natheight=431]{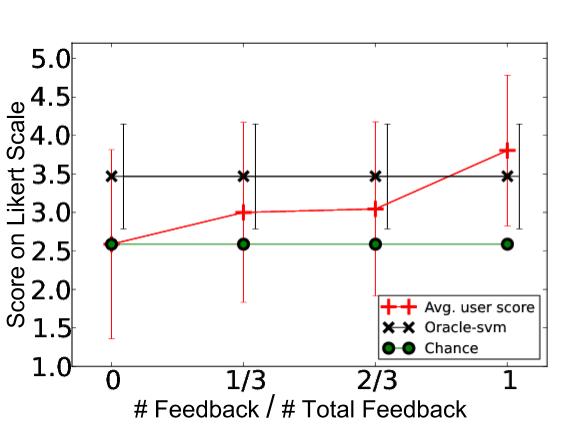}
\includegraphics[width=0.35\textwidth,height=0.26\textwidth,natwidth=571,natheight=419]{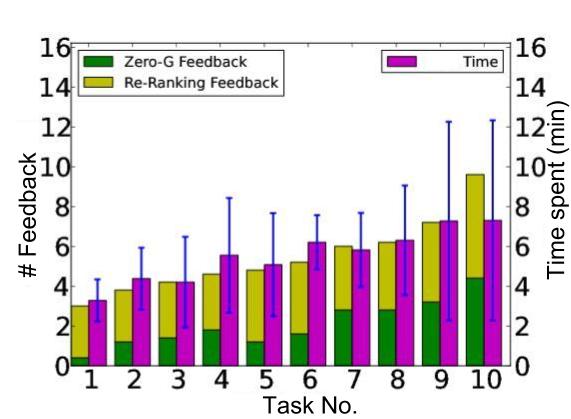}
\captionsetup{justification=centering}
\caption*{Grocery store setting on Baxter.}
\includegraphics[width=0.35\textwidth,height=0.26\textwidth,natwidth=575,natheight=445]{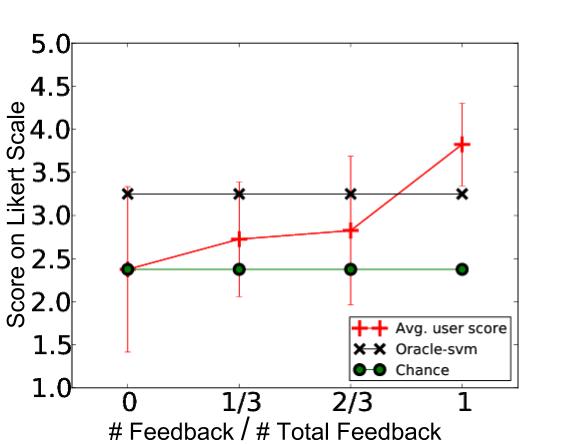}
\includegraphics[width=0.35\textwidth,height=0.26\textwidth,natwidth=574,natheight=451]{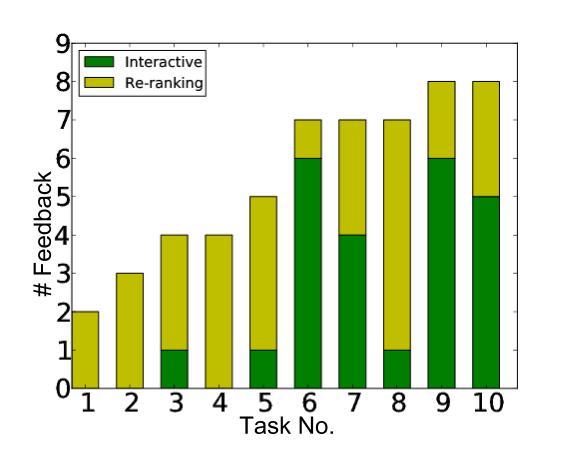}
\caption*{Household setting on PR2.}
\captionsetup{justification=justified}
\caption{\textbf{(Left)} Average quality of the learned trajectory after every one-third of total feedback. \textbf{(Right)} Bar chart showing the average number of 
feedback (re-ranking and zero-G) and time required (only for grocery store setting) for each task. Task difficulty increases from 1 to 10.}
\label{fig:userplot}
\end{figure*}

\noindent We perform a user study of our system on Baxter and PR2 on a variety of tasks of varying difficulties
in grocery store and household settings, respectively. Thereby we show a
proof-of-concept of our approach in real world robotic scenarios, and that the combination 
of re-ranking and zero-G/interactive feedback allows users to train the robot in
few feedback iterations. 
\\~\\~
\header{Experiment setup:} In this study, users not associated with this work, used our 
system to train PR2 and Baxter on household and grocery checkout tasks,
respectively. 
Five users independently trained Baxter, by providing zero-G feedback
kinesthetically on the robot, and re-rank feedback in a simulator.
Two users participated in the study on PR2. On PR2, in place of zero-G, users
provided interactive waypoint correction feedback in the Rviz simulator.
The users were undergraduate students.  
Further, both users training PR2 on household tasks were familiar with Rviz-ROS.\footnote{The 
smaller user size on PR2 is because it requires users with experience in Rviz-ROS. Further, we also 
observed users found it harder to correct trajectory waypoints in a simulator than providing zero-G feedback on the robot. 
For the same reason we report training time only on Baxter for grocery store setting.}
A set of 10 tasks of varying difficulty level was presented to users one at a time, and they were 
instructed to provide feedback until they were satisfied with the top ranked trajectory. 
To quantify the quality of learning each user evaluated their own trajectories (self score), 
the trajectories learned by the other users (cross score), and those predicted by Oracle-svm, 
on a Likert scale of 1-5. We also recorded the total time a user
spent on each 
task -- from start of training till the user was satisfied with the top ranked
trajectory. This includes time taken for both re-rank and zero-G feedback.
\\~\\~
\begin{figure*}[t]
\centering
\includegraphics[width=.75\linewidth,natwidth=1655,natheight=460]{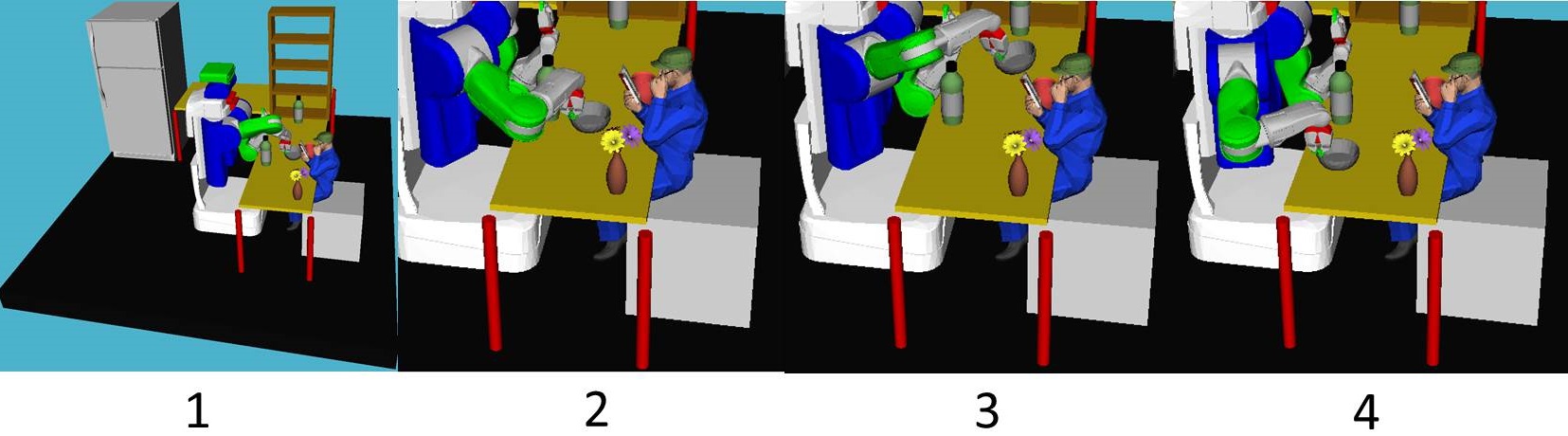}
\caption{Shows trajectories for moving a bowl of water in presence of human. Without learning robot plans an undesirable
trajectory and moves bowl over the human (waypoints 1-3-4). After six user feedback robot learns the desirable trajectory (waypoints 1-2-4).}
\label{fig:traj_1}
\end{figure*}
\begin{figure*}[t]
\centering
\includegraphics[width=0.25\linewidth,height=0.25\linewidth,natwidth=761,natheight=790]{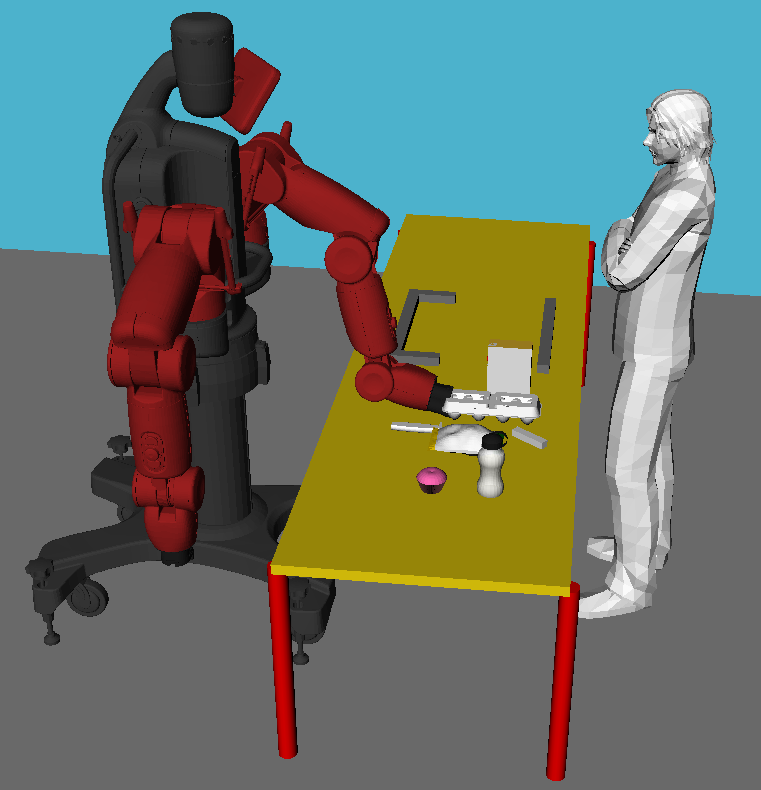}
\includegraphics[width=0.25\linewidth,height=0.25\linewidth,natwidth=849,natheight=869]{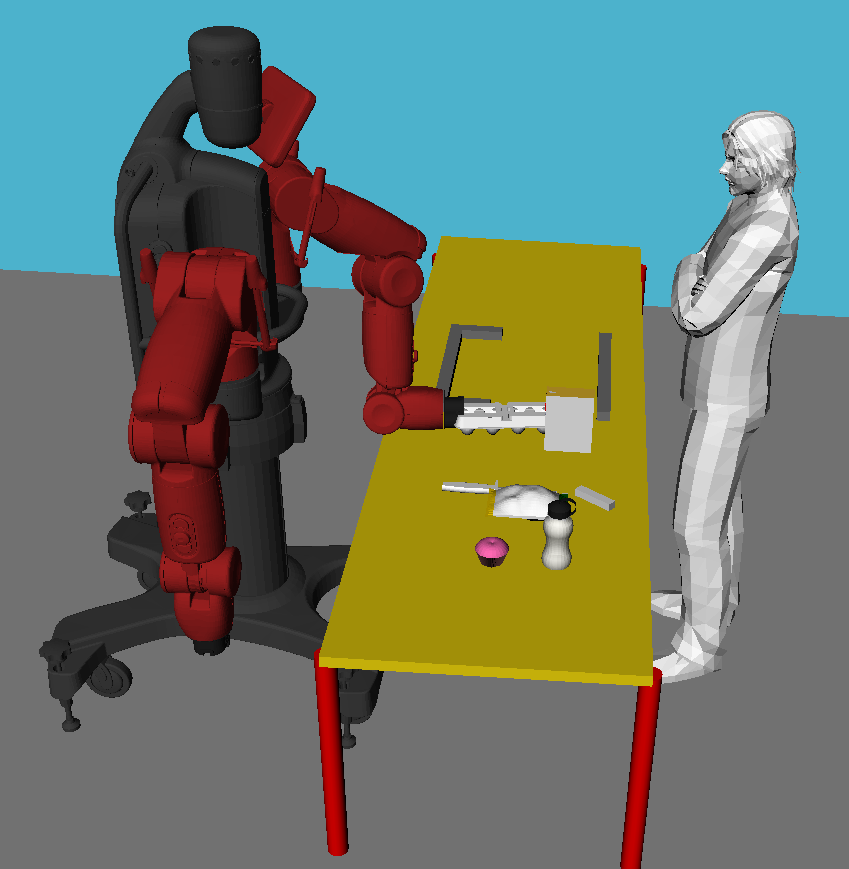}
\includegraphics[width=0.25\linewidth,height=0.25\linewidth,natwidth=889,natheight=906]{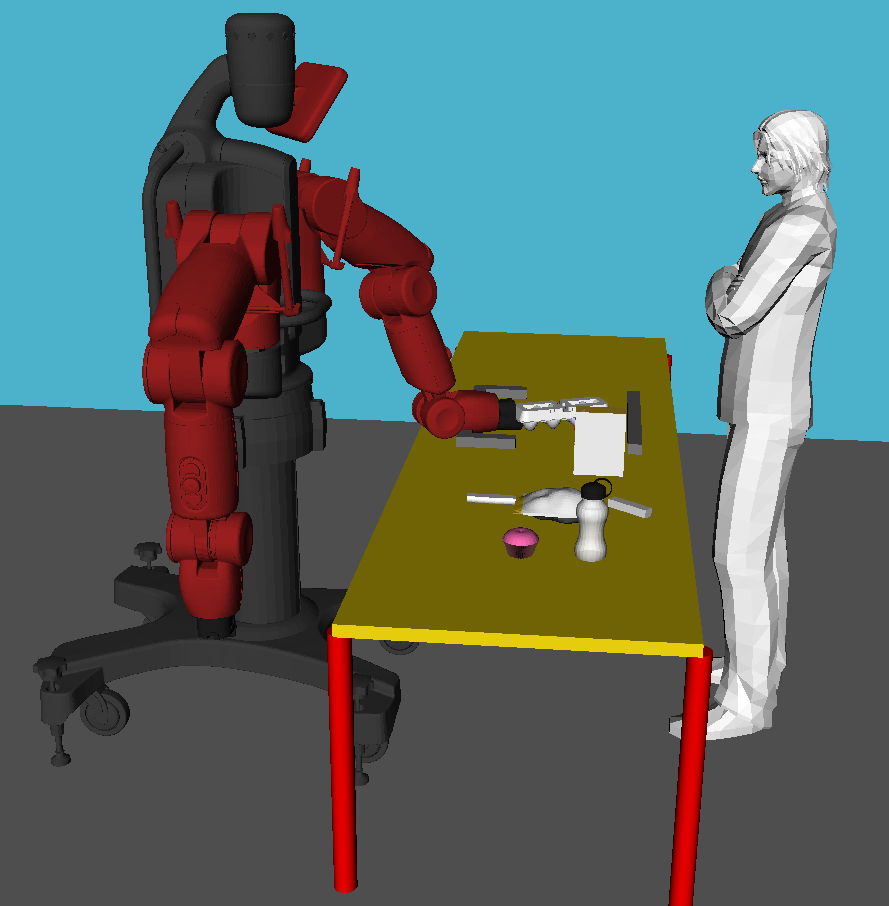}
\caption{Shows the learned trajectory for moving an egg carton. Since eggs are fragile robot moves the carton
near the table surface. (Left) Start of trajectory. (Middle) Intermediate waypoint with egg close to the table surface.
(Right) End of trajectory.}
\label{fig:traj_2}
\end{figure*}
\header{Is re-rank feedback easier to elicit from users than zero-G or interactive?} In our user study,
on average a user took 3 re-rank and 2 zero-G feedback per task to train a robot (Table~\ref{tab:userstudy}).
From this we conjecture, that for high DoF manipulators re-rank feedback is easier to provide than 
zero-G -- which requires modifying the manipulator joint angles. However, an increase in 
the count of zero-G (interactive) feedback with task difficulty suggests
(Figure~\ref{fig:userplot} (Right)), 
users rely more on zero-G feedback for difficult tasks since it allows precisely 
rectifying erroneous waypoints. Figure~\ref{fig:traj_1} and Figure~\ref{fig:traj_2} show two example trajectories learned
by a user.
\\~\\~
\header{How many feedback iterations a user takes to improve over Oracle-svm?} 
Figure~\ref{fig:userplot} (Left) shows that the quality of trajectory improves
with feedback. On average, a user took 5 feedback to improve over 
Oracle-svm, which is also consistent with our quantitative 
analysis (Section~\ref{subsec:oracle}). In grocery setting, users 4 and 5 were critical towards trajectories learned by Oracle-svm and gave them 
low scores. This indicates a possible mismatch in preferences between our expert
(on whose labels Oracle-svm was
trained) and users 4 and 5.
\begin{table}[t]
\centering
\begin{tabular}{@{}cccccc}
\hline
\multirow{2}{*}{User}&\# Re-ranking & \# Zero-G & Average & \multicolumn{2}{c}{Trajectory-Quality}\\
&feedback&feedback&time (min.)&self&cross\\\hline
1&5.4 (4.1)&3.3 (3.4)&7.8 (4.9)&3.8 (0.6)&4.0 (1.4)\\
2&1.8 (1.0)&1.7 (1.3)&4.6 (1.7)&4.3 (1.2)&3.6 (1.2)\\
3&2.9 (0.8)&2.0 (2.0)&5.0 (2.9)&4.4 (0.7)&3.2 (1.2)\\
4&3.2 (2.0)&1.5 (0.9)&5.3 (1.9)&3.0 (1.2)&3.7 (1.0)\\
5&3.6 (1.0)&1.9 (2.1)&5.0 (2.3)&3.5 (1.3)&3.3 (0.6)\\\hline
\\
\end{tabular}
\begin{tabular}{@{}ccccc}
\hline
\multirow{2}{*}{User}&\# Re-ranking & \# Interactive & \multicolumn{2}{c}{Trajectory-Quality}\\
&feedbacks&feedbacks&self&cross\\\hline
1&3.1 (1.3)&2.4 (2.4)&3.5 (1.1) & 3.6 (0.8)\\
2&2.3 (1.1)&1.8 (2.7)&4.1 (0.7) & 4.1 (0.5)\\
\hline
\end{tabular}
\caption{Shows learning statistics for each user. Self and cross scores of the final learned trajectories. 
The number inside bracket is standard deviation. \textbf{(Top)} Results for grocery store on Baxter. \textbf{(Bottom)} 
Household setting on PR2.}
\label{tab:userstudy}
\end{table}
\\~\\~
\header{How do users' unobserved score functions vary?} An average difference of 0.6 between users' self and cross score (Table~\ref{tab:userstudy})
in the grocery checkout setting suggests preferences varied across users, but only marginally. In situations 
where this difference is significant and a system is desired for a user population, a future work 
might explore coactive learning for satisfying user population, similar to Raman and Joachims~\citep{Raman13}.
For household setting, the sample size is small to draw a similar conclusion.
\\~\\~
\header{How long does it take for users to train a robot?}  We report training
time for only the grocery store setting, because
the interactive feedback in the household setting requires users with experience in Rviz-ROS. Further, we observed that users found it
difficult to modify the robot's joint angles in a simulator to their desired configuration.
In the grocery checkout setting, among all the users,
user 1 had the strictest preferences and also experienced some early difficulties in using the 
system and therefore took longer than others. On an average, a user took 5.5 minutes per 
task, which we believe is acceptable for most applications.
Future research in human-computer interaction, visualization and better user
interfaces~\citep{Shneiderman03} could 
further reduce this time. For example, simultaneous visualization of top ranked trajectories instead of  
sequentially showing them to users (the scheme we currently adopt) could bring down the time for re-rank feedback.
Despite its limited size, through our user study we show that our algorithm is
realizable in practice on high DoF manipulators. We hope this motivates researchers to build 
robotic systems capable of learning from \textit{non-expert} users.
For more details, videos and code,  visit: 

{\tt \hspace{5mm} \url{http://pr.cs.cornell.edu/coactive/}}

\section{Conclusion and Future work} 
\label{sec:conclusion}

When manipulating objects in human environments, it is important for robots to plan motions
that follow users' preferences.  In this work, we considered preferences that go beyond simple
geometric constraints and that considered surrounding context of various objects and humans
in the environment.   We presented a coactive learning approach for teaching robots 
these preferences through iterative improvements from non-expert users.
Unlike in standard learning from demonstration approaches, our approach 
does not require the user to provide optimal trajectories 
as training data.  We evaluated our approach on various household (with PR2) and grocery store 
checkout (with Baxter) settings.  Our experiments suggest that it is indeed possible
to train robots within a few minutes with just a few incremental feedbacks from non-expert users.

Future research could extend coactive learning to situations
with uncertainty in object pose and attributes. Under uncertainty the trajectory
preference perceptron will admit a belief space update form, and theoretical
guarantees will also be different. Coactive feedback might also find use in other
interesting robotic applications such as assistive cars, where a car learns
from humans steering actions. Scaling up coactive feedback by crowd-sourcing and
exploring other forms of easy-to-elicit learning signals are also potential
future directions. 

\section*{Acknowledgements} This research was supported by ARO award W911NF-12-1-0267, 
Microsoft Faculty fellowship and NSF Career award (to Saxena).

\appendices
\section{Proof for Average Regret}
\label{subsec:proof}
This proof builds upon Shivaswamy \& Joachims~\citep{Shivaswamy12}.

We assume the user hidden score function $s^*(x,{y})$ 
is contained in the family of scoring functions $s(x,{y};w_O^*,w_E^*)$ for some unknown $w_O^*$ and $w_E^*$.
Average regret for TPP over $T$ rounds of interactions can be written as:
\begin{align}
\nonumber REG_T &= \frac{1}{T} \sum_{t=1}^T ({s}^*(x_t,{{y}}_t^*)- {s}^*(x_t,{{y}}_t))\\
\nonumber &=\frac{1}{T} \sum_{t=1}^T ({s}(x_t,{{y}}_t^*;w_O^*,w_E^*)- {s}(x_t,{{y}}_t;w_O^*,w_E^*))
\end{align}

We further assume the feedback provided by the user is strictly $\alpha$-informative and satisfy following inequality:
\begin{align}
\label{eq:alphafeedback}
\nonumber s(x_t,\bar{y}_t;w_O^*,w_E^*) \geq &s(x_t,y_t;w_O^*,w_E^*)+ \alpha[{s}(x_t,{{y}}_t^*;w_O^*,w_E^*)\\ 
&- {s}(x_t,{{y}}_t;w_O^*,w_E^*)] - \xi_t
\end{align}

Later we relax this constraint and requires it to hold only in expectation.

This definition states that the user feedback should have a score of $\bar{y}_t$ that is higher than that of $y_t$ 
by a fraction $\alpha \in (0,1]$ of the maximum possible range $s(x_t,y_t^*;w_O^*,w_E^*)-s(x_t,y_t;w_O^*,w_E^*)$.
\\~\\~
\begin{thm}
 \textit{The average regret of trajectory preference perceptron receiving strictly $\alpha$-informative feedback can be
 upper bounded for any $[{w}_O^*;{w}_E^*]$ as follows:}
 \begin{equation}
\label{eq:regret-bound}
REG_T \leq \frac{2C\left\|[{w}_O^*;{w}_E^*]\right\|}{\alpha \sqrt{T}}  + \frac{1}{\alpha T}\sum_{t=1}^T \xi_t
 \end{equation}
 \textit{where C is constant such that $\left\|\left[\phi_O(x,{y});\phi_E(x,{y}) \right]\right\|_{2} \leq C$.}
\end{thm}
\textbf{Proof:} After $T$ rounds of feedback, using weight update equations of $w_E$ and $w_O$ we can write:
\begin{align}
\nonumber {w}_O^{*}\cdot{w}_O^{(T+1)} &= {w}_O^{*}\cdot{w}_O^{(T)} +  {w}_O^{*}\cdot({\phi_O}(x_T,{\bar{{y}}_T}) - {\phi_O}(x_T,{{y}_T}))\\
\nonumber {w}_E^{*}\cdot{w}_E^{(T+1)} &= {w}_E^{*}\cdot{w}_E^{(T)} +  {w}_E^{*}\cdot({\phi_E}(x_T,{\bar{{y}}_T}) - {\phi_E}(x_T,{{y}_T}))
\end{align}

Adding the two equations 
 and recursively 
reducing the right gives:
\begin{align}
\label{eq:bound-1}
\nonumber{w}_O^{*}\cdot{w}_O^{(T+1)} + {w}_E^{*}\cdot{w}_E^{(T+1)} = &\sum_{t=1}^T
({s}(x_t,{\bar{y}}_t;w_O^*,w_E^*)\\
& - {s}(x_t,{{y}}_t;w_O^*,w_E^*))
\end{align}

Using Cauchy-Schwarz inequality the left hand side of equation~\eqref{eq:bound-1} can be bounded as:
\begin{equation}
\label{eq:caucy}
{w}_O^{*}\cdot{w}_O^{(T+1)} + {w}_E^{*}\cdot{w}_E^{(T+1)} \leq
\left\|[{w}_O^*;{w}_E^*]\right\|\left\|[{w}_O^{(T+1)};{w}_E^{(T+1)}]\right\|\\  
\end{equation}
$\left\|[{w}_O^{(T+1)};{w}_E^{(T+1)}]\right\|$ can be bounded by using weight update equations:
\begin{align}
\nonumber
&{w}_O^{(T+1)}\cdot{w}_O^{(T+1)} + {w}_E^{(T+1)}\cdot{w}_E^{(T+1)} = {w}_O^{(T)}\cdot{w}_O^{(T)} + {w}_E^{(T)}\cdot{w}_E^{(T)} \\
\nonumber &+
2{w}_O^{(T)}\cdot({\phi_O}(x_T,{\bar{{y}}_T}) - {\phi_O}(x_T,{{y}_T}))\\
\nonumber
& +
2{w}_E^{(T)}\cdot({\phi_E}(x_T,{\bar{{y}}_T}) - {\phi_E}(x_T,{{y}_T}))
\\
\nonumber
&+({\phi_O}(x_T,{\bar{{y}}_T}) - {\phi_O}(x_T,{{y}_T}))\cdot({\phi_O}(x_T,{\bar{{y}}_T}) - {\phi_O}(x_T,{{y}_T}))\\
\nonumber
&+({\phi_E}(x_T,{\bar{{y}}_T}) - {\phi_E}(x_T,{{y}_T}))\cdot ({\phi_E}(x_T,{\bar{{y}}_T}) - {\phi_E}(x_T,{{y}_T}))\\
\label{eq:bound-regret}
&\leq {w}_O^{(T)}\cdot{w}_O^{(T)} + {w}_E^{(T)}\cdot{w}_E^{(T)} + 4C^2 \leq 4C^2T\\
\label{eq:bound-norm}
&\therefore \left\|[{w}_O^{(T+1)};{w}_E^{(T+1)}]\right\| \leq 2C\sqrt{T}
\end{align}

Eq. \eqref{eq:bound-regret} follows from the fact that $s(x_T,{{y}}_T;w_O^{(T)},w_E^{(T)}) > s(x_T,\bar{{y}}_T;w_O^{(T)},w_E^{(T)})$
and  $\left\|\left[\phi_O(x,{y});\phi_E(x,{y}) \right]\right\|_{2} \leq C$. Using equations~\eqref{eq:caucy} and~\eqref{eq:bound-norm} gives following bound
on~\eqref{eq:bound-1}:
\begin{align}
\label{eq:bound-2}
\nonumber \sum_{t=1}^T ({s}(x_t,{\bar{y}}_t;w_O^*,w_E^*) -
{s}(&x_t,{{y}}_t;w_O^*,w_E^*))\\
& \leq 2C\sqrt{T}\left\|[{w}_O^*;{w}_E^*]\right\|
\end{align}
Assuming strictly $\alpha$-informative feedback and re-writing equation~\eqref{eq:alphafeedback} as:
\begin{align}
\label{eq:alphafeed}
\nonumber &s(x_t,{y}_t^*;w_O^*,w_E^*) - s(x_t,y_t;w_O^*,w_E^*) \\
&\leq \frac{1}{\alpha}(({s}(x_t,{\bar{y}}_t;w_O^*,w_E^*) - {s}(x_t,{{y}}_t;w_O^*,w_E^*)) - \xi_t)
\end{align}

Combining equations~\eqref{eq:bound-2} and~\eqref{eq:alphafeed} gives the bound on average regret~\eqref{eq:regret-bound}.

\section{Proof for Expected Regret}
\label{subsec:proof-cor}
We now show the regret bounds for TPP under a weaker feedback assumption -- expected $\alpha$-informative feedback:
\begin{align}
\label{eq:expecalphafeedback}
\nonumber E_t[s(x_t,\bar{y}_t;&w_O^*,w_E^*)] \geq s(x_t,y_t;w_O^*,w_E^*) \\
\nonumber &+ \alpha[{s}(x_t,{{y}}_t^*;w_O^*,w_E^*)- {s}(x_t,{{y}}_t;w_O^*,w_E^*)] - \xi_t
\end{align}
where the expectation is under choices $\bar{y}_t$ when $y_t$ and $x_t$ are known.
\begin{cor}
 \textit{The expected regret of trajectory preference perceptron receiving expected $\alpha$-informative feedback can be
 upper bounded for any $[{w}_O^*;{w}_E^*]$ as follows:}
 \begin{equation}
\label{eq:cor-regret-bound}
E[REG_T] \leq \frac{2C\left\|[{w}_G^*;{w}_O^*]\right\|}{\alpha \sqrt{T}}  + \frac{1}{\alpha T}\sum_{t=1}^T \bar{\xi}_t
 \end{equation}
 \end{cor}
 \textbf{Proof:} Taking expectation on both sides of equation~\eqref{eq:bound-1},~\eqref{eq:caucy} and~\eqref{eq:bound-regret} yields following equations respectively:

\begin{align}
\label{eq:new-bound-1}
\nonumber E[{w}_O^{*}\cdot{w}_O^{(T+1)} + {w}_E^{*}\cdot{w}_E^{(T+1)}] &= \sum_{t=1}^T
E[({s}(x_t,{\bar{y}}_t;w_O^*,w_E^*)\\
& - {s}(x_t,{{y}}_t;w_O^*,w_E^*))]
\end{align}

\begin{align}
\nonumber E[{w}_O^{*}\cdot{w}_O^{(T+1)} + &{w}_E^{*}\cdot{w}_E^{(T+1)}]\\
\nonumber & \leq \left\|[{w}_O^*;{w}_E^*]\right\| E\left[\left\|[{w}_O^{(T+1)};{w}_E^{(T+1)}]\right\|\right]
\end{align}

\begin{flalign}
\nonumber E[{w}_O^{(T+1)}\cdot{w}_O^{(T+1)} + {w}_E^{(T+1)}\cdot{w}_E^{(T+1)}] \leq 4C^2T
\end{flalign}

Applying Jensen's inequality on the concave function $\sqrt{\cdot}$ we get:
\begin{align}
\nonumber
&E[{w}_O^{*}\cdot{w}_O^{(T+1)} + {w}_E^{*}\cdot{w}_E^{(T+1)}] \\
\nonumber &\leq \left\|[{w}_O^*;{w}_E^*]\right\|E\left[\left\|[{w}_O^{(T+1)};{w}_E^{(T+1)}]\right\|\right]\\
\nonumber 
&\leq \left\|[{w}_O^*;{w}_E^*]\right\|\sqrt{E[{w}_O^{(T+1)}\cdot{w}_O^{(T+1)} + {w}_E^{(T+1)}\cdot{w}_E^{(T+1)}]}
\end{align}
Using \eqref{eq:new-bound-1} gives the following bound:
\begin{align}
\nonumber
\sum_{t=1}^T E[{s}(x_t,{\bar{y}}_t;w_O^*,w_E^*) &-
{s}(x_t,{{y}}_t;w_O^*,w_E^*)] \\
\nonumber & \leq 2C\sqrt{T}\left\|[{w}_O^*;{w}_E^*]\right\|
\end{align}
Now using the fact that the user feedback is expected $\alpha$-informative gives the regret bound \eqref{eq:cor-regret-bound}.

\end{document}